\documentclass[11pt]{article}
\usepackage[preprint]{acl}   

\usepackage{times}
\usepackage{latexsym}
\usepackage[T1]{fontenc}
\usepackage[utf8]{inputenc}
\usepackage{microtype}
\usepackage{inconsolata}

\usepackage{graphicx}
\usepackage{amsmath}
\usepackage{amssymb}
\usepackage{booktabs}
\usepackage{mdframed}
\usepackage{float}
\usepackage{capt-of}

\title{How Relations Impose Embedding-Space Structure}

\author{
  \textbf{Lachlan Cowley\textsuperscript{1}},
  \textbf{Jey Han Lau\textsuperscript{2}},
  \textbf{Yulia Otmakhova\textsuperscript{2}},
  \textbf{Eduard Hovy\textsuperscript{2}}
  \\
  \textsuperscript{1}Monash University,
  \textsuperscript{2}The University of Melbourne
  \\
  \small{\textbf{Correspondence:} \texttt{lcow0010@student.monash.edu}}
}
\begin{document}
\maketitle

\begin{abstract}

Language models are known to encode simple relations, such as synonymy, in the geometry of their representations. It is unclear whether they also encode complex patterns of relationships such as those that drive analogy. We present a simple method, with no learned parameters, based on comparing the internal structural arrangements between disparate domains, to test this. Using our method we find over multiple models that representations are arranged consistently to encode complex relationship similarities. Models align distant representation clouds so that facets of meaning are easily comparable under simple mappings. 
Our work extends mechanistic accounts of representation space to the relative arrangement of representations that may be only weakly related to one another, showing that representation geometry is far richer than previously thought\footnote{Code and data: \url{https://github.com/lcowley7/embedded_relationships}}.
\end{abstract}

\begin{figure*}[t]
    \centering
    \includegraphics[width=0.75\linewidth]{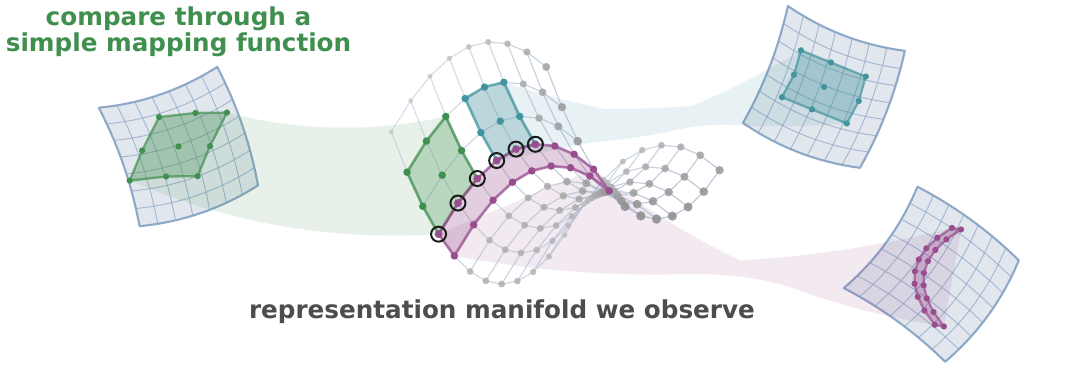}
    \caption{ An illustration of how accumulated relationships could shape a domain's geometry. We hypothesise that each domain must remain comparable, through a simple map, with parts of other, distant domains it is related to, and that these requirements fix how its members are arranged. Requirements from different domains overlap on some members (ringed), so no flat arrangement satisfies them all; here, the arrangement that best satisfies them is saddle-shaped. We are testing this via whether distant but related manifold pairs are nearly second-order isomorphic, which gives us an estimate of how simple a map between the two spaces could be. Overlapping requirements of this kind could explain the extra curvature observed in representation manifolds, which bend through more dimensions than their own structure requires \citep{olah2024multidimensional, modell2025origins}.}

    \label{fig:diagram_alignment}
\end{figure*}

\section{Introduction}

Instead of defining a word intensionally as a collection of features (the {\em differentiae} of Aristotle) or extensionally as the set of all examples of it, as done by \citet{frege1892sinn, carnap1947meaning}, modern distributional approaches follow Firth’s famous dictum ``you shall know a word by the company it keeps” \citep{firth1957synopsis}. In the distributional paradigm the meaning of a word or a concept is captured by the set of relationships it has with other words or concepts.

Words participate in a dense web of complex lexical and semantic relationships. It is well established that synonymy and hypernymy are encoded in language model representation geometry \citep{mikolov_linguistic_nodate, parkGeometryCategoricalHierarchical2025}. But what about less straightforward relationships, such as the connection between \textit{boiling water}${\leftrightarrow}100\,^{\circ}$C, or, in the \textsc{argument is war} metaphor \citep{lakoff2008metaphors}, the links between \textit{attack}$\leftrightarrow$\textit{criticise} or \textit{defend}$\leftrightarrow$\textit{justify}?

In this paper, we explore how modern large language models (LLMs) organise their embedding representation space: how words are defined
by the relationships they have with other words. We explore how models encode information about the world as they know it in the arrangement of their representations. Specifically,
we look at analogies between pairs of distant domains (e.g., water-state words vs.\ temperature words) and investigate whether LLMs arrange the representations of their concepts (e.g., \textit{boiling water}) to encode these complex, distant relationships. 

We are interested in distant domains because they presumably have a distinct local topology in the embedding space \citep{engels2024not}. However, if the model uses geometry to encode cross-domain analogies, the domains should exhibit some shared structure. Understanding how the model balances internal local structure and cross-domain analogical structure, and whether it encodes relationships, is important for understanding how the model uses its representation geometry in general, a question that impacts interpretability and monitoring.

We first illustrate our method with water state words and Celsius temperatures. Despite being far apart in representation space, a model with a good understanding of the world may relate the two domains (such as \textit{icy water}${\leftrightarrow}1\,^{\circ}$C and \textit{boiling water}${\leftrightarrow}100\,^{\circ}$C). It may do so through 
an explicit shared direction for each link
or through two parallel
local structures,
that are similar to each other. The former approach would populate the space with a vast number of directions that encode cross-domain analogy links. We determine in this paper that instead LLMs prefer the latter approach. 

To do so we first obtain `pure' concept representations by creating steering vectors \citep{wu2025axbench}: vectors that, when added during generation, orient the model to the desired concept. To obtain the structure of a domain, we capture how domain concepts change as temperature changes. We call the difference between two concepts in the same domain a \textbf{move}: for example, \textit{icy water}$\rightarrow$\textit{boiling water}, which matches its counterpart $1{\rightarrow}100\,^{\circ}$C. Within each domain we measure the similarity of every move to every other move, and then determine how closely the two domains agree on the overall pattern of moves. If \textit{icy water}$\rightarrow$\textit{boiling water} resembles \textit{cold water}$\rightarrow$\textit{hot water} more than \textit{cool water}$\rightarrow$\textit{warm water}, is $1{\rightarrow}100\,^{\circ}$C likewise closer to $8{\rightarrow}55\,^{\circ}$C than to $15{\rightarrow}35\,^{\circ}$C? Agreement of this kind is what \citet{shepardSecondorderIsomorphismInternal1970} called a \emph{second-order isomorphism}: the domains need not share vector directions in the representation space, only the pattern of relations among their members. It is the minimum a model would need in order to compare temperature across the two domains through a simple map.

Using our method we find that LLMs arrange their representations to encode relational patterns across domains despite conceptual distance. We find this by the consistent arrangement of temperature in each domain rather than common directions between representations of the phases of water and Celsius temperature values
. We find the same kind of correlation between analogical domains: the arrangement of representations for arguments is aligned with the arrangement of representations for warfare, and likewise for ideas and food, and importance and size. We find this arrangement alignment without explicitly needing to specify contexts that would elicit a comparison between each space (i.e., when creating the steering vector for a water word, the sentences we use do not talk about temperature). In other words, we find that models \textbf{record relevant relational knowledge through the arrangement of the representations that may take part in a comparison, even without explicit cross-domain linkage.}

\section{Related Work}

\paragraph{Linear representations} The Linear Representation Hypothesis (LRH) holds that concepts are encoded as directions in representation space. It explains why linear probes succeed \citep{alain2016understanding}, why steering vectors work \citep{turner2023activation, rimsky_steering_2024}, and why sparse autoencoders (SAEs) can recover interpretable features \citep{bricken2023monosemanticity, templeton2024scaling}. It says less about how those directions are arranged relative to one another, especially across analogical domains. Independent features are packed almost orthogonally, correlated features are arranged to interfere positively \citep{elhage2022toy}, and among nearby representations cosine distance tracks semantic distance \citep{modell2025origins}. Whether analogically similar but distant features are arranged in similar ways is largely open. \citet{gurnee2025manifolds} give one example where they are: attention heads align two distant manifolds in their query--key subspace before comparing them. We ask whether such alignment is frequent.

\paragraph{Second-order isomorphism} \citet{shepardSecondorderIsomorphismInternal1970} argue that a representation need not resemble what it represents: the mental representation of a square need not itself be square, but should be `closer' to that of a rectangle than to that of a green flash. They call such correspondence between relations a \emph{second-order isomorphism}. The idea underlies Representational Similarity Analysis (RSA) \citep{kriegeskorte2008representational}, which is now used in NLP \citep{abnar2019blackbox, chrupala2019correlating, lepori2020picking} and which we adopt. For two domains, a first-order correspondence between two concepts would be captured by a shared representation vector direction, as the LRH would suggest. Reaching across distant domains, we test for the second-order one, extending mechanistic explanations of representations from vector directions to their arrangement.

\paragraph{Curved manifolds} \citet{olah2024multidimensional} and \citet{modell2025origins} note that representation manifolds in LLMs exhibit more curvature and are higher dimensional than appears necessary to just convey the qualities of the feature they encode. We seek to understand if that curvature may (in part) be due to models arranging many related but distant manifolds from different domains to be easily comparable under simple mappings. Figure~\ref{fig:diagram_alignment} shows our intuition.

\paragraph{Comparing distant domains} One option for comparing domains is to fit an explicit mapping between two spaces \citep{hewitt2019structural}. However, in high dimensions fitted mappings can find structure that is not there \citep{ravichander2021probing}. 

Centered Kernel Alignment \citep[CKA]{kornblith2019similarity} compares similarity using vector magnitudes and directions. However, absolute cosine distance diverges from semantic distance between far apart concepts \citep{modell2025origins}. But since their rank order remains \citep{modell2025origins}, we compare only rank orders. We do so by appropriate operations on a distance matrix that achieve statistical significance \citep{mantel1967detection}.

\section{Method}
\label{sec:method}

\begin{figure*}[t]
    \centering
    \includegraphics[width=\linewidth]{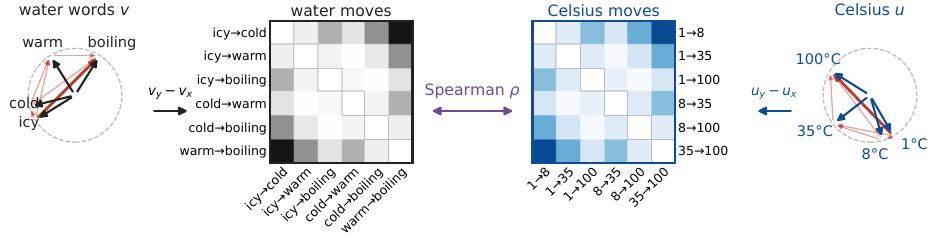}
    \caption{Our method on a toy example in two dimensions. \textbf{Outer panels:} steering vectors for four water state words ($v$) and four Celsius temperatures ($u$). Red arrows are \emph{moves}, the differences between two concepts in the same domain, from colder to warmer; the bold arrow marks \textit{icy water}$\rightarrow$\textit{boiling water} and its counterpart $1{\rightarrow}100\,^{\circ}$C. \textbf{Inner panels:} within each domain, the cosine distance between every pair of moves. \textbf{Middle:} agreement is the Spearman $\rho$ between the two matrices over the $\binom{6}{2}=15$ pairs of moves (which is just the upper right triangle of the distance matrices).}
    \label{fig:method}
\end{figure*}

To uncover and verify cross-domain analogical structure, our method adopts Representational Similarity Analysis (RSA) \citep{kriegeskorte2008representational}, an established technique in neuroscience and NLP \citep{abnar2019blackbox, chrupala2019correlating, lepori2020picking}. Our novel contribution is a method of applying RSA using high-quality steering vectors and rank correlation only. 
Our method uses \textbf{moves} to isolate between-element relationships within a domain, a distance matrix for each domain that captures all pairwise move distances, and a cross-domain matrix comparison using rank ordering rather than actual distance to overcome scaling differences in LLM representation space. 
If the matrices are similar (which is unlikely under random chance), we may conclude that there exists a structural similarity between the two domains, and that the model has encoded the relationship in the arrangement of its representations.
Figure~\ref{fig:method} shows our method on a toy
example.

\subsection{Steering Vectors} 
For best results it is necessary to remove spurious connotations and idiosyncratic variability from the embedding representations of the domain elements. For each concept we construct steering vectors from $72$ contrastive prompts. 

We use difference-of-means steering vectors, following the notation and implementation of \citet{wu2025axbench}. Let $h \in \mathbb{R}^{d}$ denote a residual-stream activation at the chosen layer and token position. For a set $P$ of activations on contexts that express a concept and a set $N$ of contexts that do not:

\begin{equation}
  \tilde{v} \;=\; \frac{1}{\lvert P\rvert}\sum_{h\in P} h \;-\;
          \frac{1}{\lvert N\rvert}\sum_{h\in N} h,
  \qquad
  v \leftarrow \frac{\tilde{v}}{\lVert \tilde{v}\rVert}.
  \label{eq:diffmean}
\end{equation}
If adding $\alpha \cdot v$ (where $\alpha$ is some scalar strength of the intervention) to the residual stream at a chosen layer induces the desired concept to appear in ongoing LLM output generation, we can be confident we have found an accurate representation. Steering vectors are created from activations averaged across all tokens (the longest output was 169 tokens), and for evaluation steering vectors are applied to all tokens (both performed as in AxBench \citep{wu2025axbench}). We perform our analysis (and steering) on four approximately evenly spaced layers (in \texttt{gemma-2-2b} layers: $5$, $10$, $15$, $20$). For temperature experiments, we also test additional earlier layers (results are in Appendix~\ref{app:temp_other_layers}). 

\subsection{Moves}

Since each domain element is rich in definitional facets, a \emph{move} between (the steering vectors of) two domain elements $x$ and $y$ focuses the method on the information relevant to their shared relationship (where $v_x$ is the steering vector for $x$ and $v_y$ for $y$):
\[
  m_{x \to y} \;=\; v_y - v_x ,
\] 
where by convention we order $x$ and $y$ to avoid double-counting. In the subtraction, vector components shared equally by $v_x$ and $v_y$ cancel, so a move reflects just what changes between the two elements.

\subsection{Comparing two domains}

\paragraph{Pairing the domains} 
To compare the structure of the two domains we must decide on a correspondence. Calculating all possible pairings of moves across two domains is beyond our computational budget. We therefore define the correspondence we would expect if the LLM had a robust understanding of the world and demonstrate that this mapping leads to high alignment between the two spaces, and that random mappings cannot achieve similar alignment. In the temperature example we assign every element in each domain a nominal temperature index $T(x)$ and link moves by matched temperature: the water state move \textit{icy water}$\rightarrow$\textit{boiling water} and the Celsius move $1{\rightarrow}100\,^{\circ}$C share the index $(1, 100)$.

\paragraph{Move-pair matrices} Within each domain we compute the cosine distance between every pair of moves, giving one matrix per domain:
\[
  D_A\bigl[(i,j),(k,l)\bigr] \;=\;
  1 - \cos\bigl(m^{A}_{i \to j},\, m^{A}_{k \to l}\bigr),
\]
and likewise $D_B$. Every cosine is taken within a single domain only; we never measure the angle between vectors from different domains.

\paragraph{Agreement} There are many ways to compare matrices \citep{mantel1967detection}. We measure agreement using Spearman's $\rho$ between the upper triangle entries of $D_A$ and $D_B$. Because $\rho$ uses ranks only, it requires no parameters and is unaffected by any monotone distortion of either domain's internal distances, so it requires only that rank order of cosine distance tracks semantic distance within a domain \citep{modell2025origins}.

\paragraph{Baseline} To distinguish between real and spurious correlations, we randomly shuffle the labels within the distance matrix (e.g., \textit{boiling water} may now be \textit{snow}). Then we rebuild the second domain's move distance matrix and again measure correlation. Many shuffles give a distribution of correlations achieved by similar but wrong relationships. This distribution is the \textbf{permutation null}. We report the probability $p$ that this null distribution is better than the correlation measured with the real relationship. We consider a relationship encoded in the residual stream if $p$ $<0.05$.

\subsection{Data and evaluation}

Contexts are generated by an LLM (\texttt{gemini-3.5-flash}) 
across diverse domains, and a sample of ten continuations per prompt pool is checked manually;

example pairs, generation prompts, and auditing process are in Appendix~\ref{app:sv-data}. Steering success is scored in two ways: by an LLM judge (\texttt{gemini-3.1-flash-lite}) using an adapted version of the AxBench rubrics \citep{wu2025axbench}, and, for verifiable concepts such as temperatures, by checking directly for the target term with a soft margin (rubrics in Appendix~\ref{app:rubrics}). Steering is evaluated on held-out data. We report the success of every steering vector in Appendix~\ref{app:sv-perf} and exclude none, since performance is generally strong; Appendix~\ref{app:sensitive} shows that removing the weakest vectors does not change our conclusions.
 
\section{Temperature in detail}
\label{sec:temperature}

As a guiding question we ask: Are the water phases and Celsius temperature domains arranged so that it is easy for a model to query the temperature of a phase of water?
Since temperature is only a small part of the semantics of water, and water moves are near orthogonal to Celsius moves (Table~\ref{tab:edge_temp_rsa}),
the answer is not clear.

\subsection{Setup}

We train steering vectors on \texttt{gemma-2-2b-it} for $131$ Celsius levels ($-20$ to $110\,^{\circ}$C, one per degree), $251$ Fahrenheit levels ($-10$ to $240\,^{\circ}$F), and a set of $18$ water words in Table~\ref{tab:water_words} at layers $\{0,1,2,3,4,5,10,15,20\}$.
 
We verified that no water words appeared in the numeric temperature training text, and that water-word contexts also contained no numeric temperature references. Examples are in Appendix~\ref{app:sv-data}. Three groups of words share a nominal temperature (hail/snow, icy water/slush, dew/fog/rain), and the resulting ties receive equal rank.

\begin{table}[t]
\centering
\small
\setlength{\tabcolsep}{4pt}
\begin{tabular}{@{}lp{0.74\columnwidth}@{}}
\toprule
Ladder & Words, colder to warmer (nominal $^{\circ}$C) \\
\midrule
Phase         & ice $-10$, slush $1$, dew$^{\ast}$ $10$, fog$^{\ast}$ $10$, puddle$^{\S}$ $12$, steam$^{\dagger}$ $105$ \\
\addlinespace
Precipitation & hail$^{\ast}$ $-5$, snow$^{\ast}$ $-5$, frost $-3$, sleet $0$, rain $10$, steam$^{\dagger}$ $105$ \\
\addlinespace
Adjectival    & icy $1$, cold $8$, cool $15$, room-temperature $21$, warm $35$, hot $55$, boiling $100$ (each \emph{\ldots\ water}) \\
\bottomrule
\end{tabular}
\caption{The $18$ water words on the three ladders whose moves we compare. Each word's nominal temperature indexes its counterpart on the numeric ladders; the Fahrenheit index is $\tfrac{9}{5}\,^{\circ}\mathrm{C} + 32$, rounded to the $1\,^{\circ}$F grid. Moves are taken only within a ladder ($15 + 15 + 21$ pairs). $^{\ast}$Words on a ladder with the same temperature have no temperature move, leaving $49$ moves. $^{\dagger}$\emph{steam} ends both noun ladders and is one vector. $^{\S}$The liquid-water rung is the \emph{puddle} vector. The adjectival words name a temperature explicitly; the nouns do not.}
\label{tab:water_words}
\end{table}

\paragraph{Pooling} We place all $18$ words on three ordered ladders, each ordered by the temperatures in Table~\ref{tab:water_words}. We compute moves within each ladder but then create our distance matrices as the pool over all moves from all ladders. The index of each distance matrix is:
\begin{equation}
\begin{split}
     M = \bigl\{\, m_{x \to y} = v_y - v_x \  \mid \
     x, y \text{ share a ladder}, \\ T(x) < T(y) \bigr\}\,,
  \label{eq:temp-move-set}
\end{split}
\end{equation}
where $T(x)$ is the numeric temperature of $x$. $\lvert M \rvert = \binom{6}{2} + \binom{6}{2} + \binom{7}{2} - 2 = 49$ moves. The requirement $T(x) < T(y)$ orients every move from colder to warmer, so no move is recorded twice and all moves have the same sign. It also leaves out the 2 pairs on a ladder that share a nominal temperature (dew/fog and hail/snow), which have no temperature move. We require a large pooled distance matrix to provide strong statistical resolution on the permutation test. A permutation $p$ can never be smaller than one over the number of distinct reassignments. A six-word ladder allows only $6! = 720$ reassignments, so the phase or precipitation ladder alone could not report a $p$ below $\tfrac{1}{720} \approx 0.0014$. The pooled $18$ words allow more reassignments than the $5000$ permutation draws, and provide the statistical power needed to observe relationships through noise. A downside of this is that a cross-ladder pair differs in some water semantics that are not about temperature. So, \textbf{pooling makes our correlation estimate conservative}.

\begin{table}[ht]
\centering
\setlength{\tabcolsep}{4pt}
\small
\begin{tabular}{@{}lrr@{}}
\toprule
 & $^{\circ}$C & $^{\circ}$F \\
\midrule
\multicolumn{3}{@{}l}{\emph{Arrangement} (Spearman $\rho$)}\\
\quad water vs.\ temperature moves   & $+0.62$  & $+0.62$  \\
\quad null, mean                     & $+0.31$  & $+0.30$  \\
\quad null, 95th pct.                & $+0.41$  & $+0.43$  \\
\quad $p$                            & $\le 0.0002$ & $0.0010$ \\
\midrule
\multicolumn{3}{@{}l}{\emph{Direction} (mean $\cos$)}\\
\quad water vs.\ temperature moves   & $+0.07$  & $+0.08$  \\
\quad within water moves             & \multicolumn{2}{c}{$+0.16$} \\
\quad within temperature moves       & $+0.24$  & $+0.17$  \\
\bottomrule
\end{tabular}
\caption{Layer 10 of \texttt{gemma-2-2b-it}. $18$ water words drawn from three ladders, $49$ moves oriented from colder to warmer. \textbf{Arrangement:} Spearman $\rho$ between the two move distance matrices, against a permutation null. The null is centered near $+0.30$ because two moves sharing an endpoint are similar. \textbf{Direction:} a water move and its numeric counterpart have low cosine, no space has a single ``warmer'' direction.}
\label{tab:edge_temp_rsa}
\end{table}

\subsection{Results}

Water and numeric moves agree well above chance (Table~\ref{tab:edge_temp_rsa} top) at layer 10: $\rho = 0.62$ in both Celsius and Fahrenheit, against nulls centred at $0.31$ and $0.30$. Agreement exceeds the null at every layer we test, and is strongest at layers $4$--$5$ (Appendix~\ref{app:temp_other_layers}).

\paragraph{Shape, not direction} Corresponding moves are nearly orthogonal across domains: a water move and its numeric counterpart have mean cosine similarity of $0.07$ in Celsius and $0.08$ in Fahrenheit (Table~\ref{tab:edge_temp_rsa} bottom), and below $0.09$ at every layer (Appendix Table~\ref{tab:app_temp_direction_layers}). The agreement is therefore not mainly carried by one temperature direction that both domains share. It lies in {\em how the moves are arranged relative to one another}: the two domains are partially second-order isomorphic. Within each domain, moves are only weakly aligned with each other (mean cosine $0.16$--$0.24$), so neither domain has a single dominant ``warmer'' direction. This is likely because these manifolds 
curve. 
A natural conclusion is that this is how the model generalises along its geometry. It arranges these distinct representations to encode their shared temperature-ness, and to enable easy comparison by a simple mapping when needed. It is probable that many other relationships (i.e., beyond temperature) constrain the arrangement of phases of water too (as we hypothesise in Figure~\ref{fig:diagram_alignment}). 

Perhaps the relationship between \textit{boiling water} and $100\,^{\circ}$C is explained by atomic features that they have (e.g., they both share a feature that expresses ``very hot temperature'')? 
To test this we examined the SAE features active on each concept. 
We encoded the same training text with the GemmaScope SAEs \citep{lieberum_gemma_2024}. Corresponding concepts share almost no features: \emph{boiling water} and $100\,^{\circ}$C have none of their top-$20$ features in common. Appendix~\ref{app:sae} gives the full experiment.

\section{Analogies}
\label{sec:analogies}

We now consider four simple analogies that are sufficiently common that the model would have faced some training pressure towards somehow encoding them for generalisation and easy inference. We test whether the arrangement of representations on one side of the analogical mapping predicts the arrangement of representations on the other side: whether the model has encoded the analogy in its arrangement of representations. We use the  method of Sections~\ref{sec:method} and ~\ref{sec:temperature}.

\subsection{Setup}

We test three metaphors: \textsc{argument is war}, \textsc{ideas are food}, and \textsc{importance is size} \citep{lakoff2008metaphors}. Each metaphor is a list of source--target analogical mappings, for example \emph{attack}$\leftrightarrow$\emph{criticise} and \emph{digest}$\leftrightarrow$\emph{comprehend} (Table~\ref{tab:analogy_pairs}). We also test a structural analogy: five professions that have the same six relational predicates \citep{gentner1983structure} instantiated by roles. Any two professions can be compared role by role: for instance, the doctor, patient and clinic of \textsc{medicine} match the lawyer, client and law firm of \textsc{law}. Comparing each profession to the others gives $10$ pairs (Table~\ref{tab:frames_grid}).\footnote{We found that the profession frames steer poorly at layer $5$ (Appendix~\ref{app:sv-perf}).}

\begin{table*}[t]
\centering
\small
\setlength{\tabcolsep}{4pt}
\begin{tabular}{@{}r ll c ll c ll@{}}
\toprule
 & \multicolumn{2}{c}{\textsc{importance is size}} & & \multicolumn{2}{c}{\textsc{argument is war}} & & \multicolumn{2}{c}{\textsc{ideas are food}} \\
\cmidrule(r){2-3}\cmidrule(lr){5-6}\cmidrule(l){8-9}
\# & size & importance & & war & argument & & food & ideas \\
\midrule
1  & microscopic  & trivial     & & attack      & criticise          & & cook        & devise \\
2  & tiny         & negligible  & & defend      & justify            & & swallow     & accept \\
3  & small        & minor       & & ally        & supporter          & & digest      & comprehend \\
4  & medium-sized & middling    & & strategy    & reasoning          & & chew        & ponder \\
5  & sizable      & noteworthy  & & retreat     & back down          & & taste       & consider \\
6  & large        & significant & & surrender   & concede            & & spoon-feed  & oversimplify \\
7  & huge         & important   & & shoot down  & rebut              & & regurgitate & recite \\
8  & enormous     & crucial     & & demolish    & refute             & & recipe      & method \\
9  & colossal     & momentous   & & weapon      & evidence           & & ingredient  & component \\
10 &              &             & & stronghold  & conviction         & & stale       & unoriginal \\
11 &              &             & & ceasefire   & agree to disagree  & & fresh       & novel \\
12 &              &             & &             &                    & & appetite    & curiosity \\
\bottomrule
\end{tabular}
\caption{The three metaphor mappings. Each row is a pair; these pairs are what the permutation null reshuffles. Every concept is its own steering vector ($9 + 11 + 12 = 32$ pairs, $64$ vectors). 
}
\label{tab:analogy_pairs}
\end{table*}

\begin{table*}[t]
\centering
\small
\setlength{\tabcolsep}{5pt}
\begin{tabular}{@{}rclllll@{}}
\toprule
ROLE && medical & legal & dentistry & real estate & teaching \\
\midrule
PRACTITIONER && doctor       & lawyer         & dentist        & real-estate agent & teacher \\
RECIPIENT   && patient      & client         & dental patient & vendor            & student \\
VENUE       && clinic       & law firm       & dental surgery & agency            & school \\
MATTER      && illness      & lawsuit        & toothache      & property sale     & lesson \\
INTERVENTION && treatment    & representation & filling        & auction           & instruction \\
ARTIFACT     && prescription & contract       & dental chart   & listing           & report card \\
\bottomrule
\end{tabular}
\caption{The profession frames: six shared roles filled by five professions ($30$ vectors). Every domain profession is comparable to another role by role: $\binom{5}{2} = 10$ pairs of $15$ moves each. With six roles the null admits only $6! = 720$ relabellings, so no profession pair comparison can report $p$ below $0.0014$.}
\label{tab:frames_grid}
\end{table*}

Each concept is trained in its literal sense: \emph{attack} is an armed assault, and \emph{digest} is the body breaking down food. Phrases that could conflate the two domains, such as ``attack her argument'' or ``a half-baked idea'', never appear in training text. We do not ever elicit metaphorical comparisons; any measured correlation must be due to analogical meaning that is latently encoded (through arrangement) as part of the generic representation.

As previously, the permutation null reassigns which target is paired with which source. We form a move between every pair of concepts in a domain ($36$, $55$, $66$ and $15$ for the importance, argument, ideas, and profession domains respectively). 
\begin{figure}[t]
    \centering
    \includegraphics[width=\linewidth]{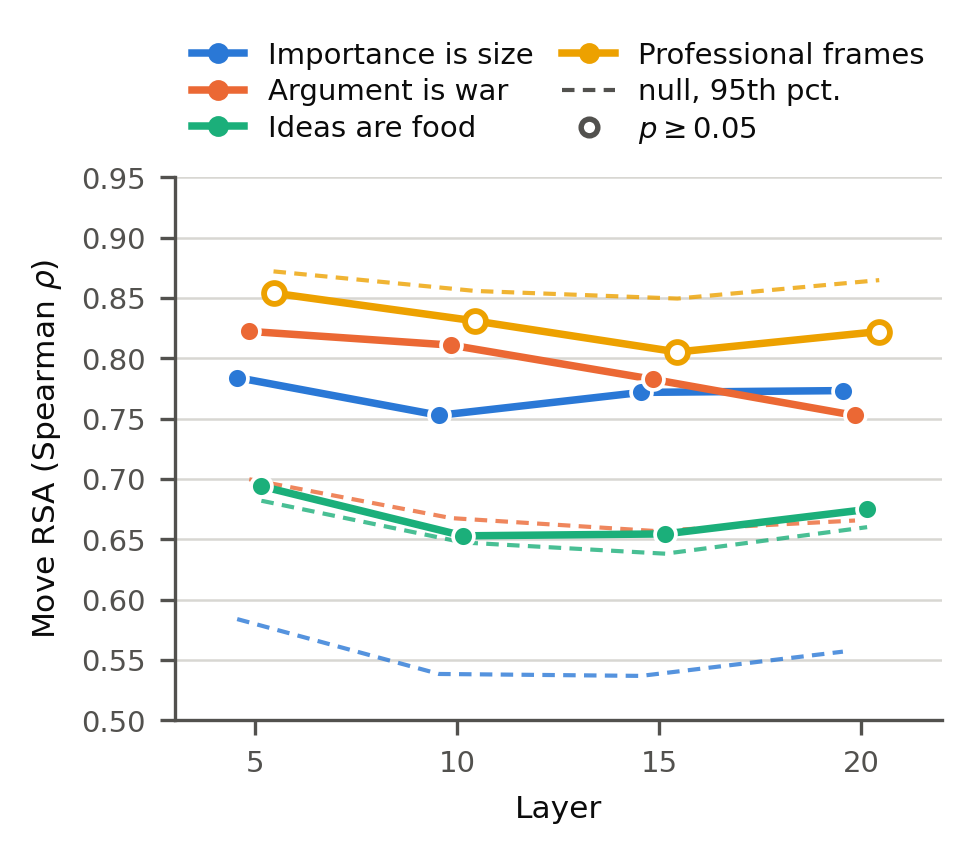}
    \caption{Move agreement between the two sides of each analogy across depth. Dashed lines are each domain's permutation-null $95$th percentile; they are above zero ($0.54$--$0.87$) because two moves that share an endpoint are similar whatever the pairing. Filled markers clear the null ($p<0.05$), hollow markers do not. For profession frames the solid line is the mean over the $10$ profession pairs, the dashed line the mean of their nulls, and a marker is filled only if most pairs clear.}
    \label{fig:analogy_rsa}
\end{figure}

\subsection{Results}

Figure~\ref{fig:analogy_rsa} shows the results, which are split by domain.
Similar to temperature, the metaphor representations are aligned. The analogy is likely a small part of each concept's semantics but is still arranged consistently. As with temperature, corresponding moves point in almost unrelated directions (mean cosine similarity is $0.07$--$0.25$; Appendix~\ref{app:analogies}). Sensitivity analysis in Appendix~\ref{app:sensitive} shows that \textsc{ideas are food} is sensitive to the removal of low efficacy steering vectors. Although $\rho$ rises, the permutation null does too, and it is not significant at layer $10$ once its single non-steering pair is removed. We think this is primarily due to our limited statistical power.

\paragraph{The profession frames do not clearly align} Only $2$ of $10$ profession pairs clear their own null (Table~\ref{tab:app_frames_pairs}). 
We believe this may be because the professions are not distant domains. When we compute the cosine similarity of the concept steering vectors (i.e., not the move vectors) between domains (Appendix Tables~\ref{tab:app_analogy_direction} and~\ref{tab:app_frames_pairs}), our suspicion is borne out: on average, a profession's vectors are almost as similar to another profession's (cosine $0.70$--$0.78$) as to each other ($0.84$--$0.87$). By contrast, the other metaphor domains are further apart ($0.37$--$0.64$). We think this means that moves do not cancel out all unrelated structure, and the similarity induced by the professional frame analogy is overwhelmed by other relations. It is also possible however that models treat these professional pairs differently to our other pairs of domains.

\section{Verification using other models}
\label{verification}

We also validate our results on \texttt{OLMo-3.1-32B-Instruct} \citep{olmoOlmo32026} and \texttt{Qwen3-32B} \citep{yangQwen3TechnicalReport2025}. For both models we compute steering vectors from the same training text as before. We compare layers at the same relative depth: layer $L$ of Gemma's $26$ is matched with layer $\operatorname{round}(64L/26)$ of the $64$-layer models. Appendix~\ref{app:models} shows the details.
\begin{figure*}[t]
    \centering
    \includegraphics[width=\linewidth]{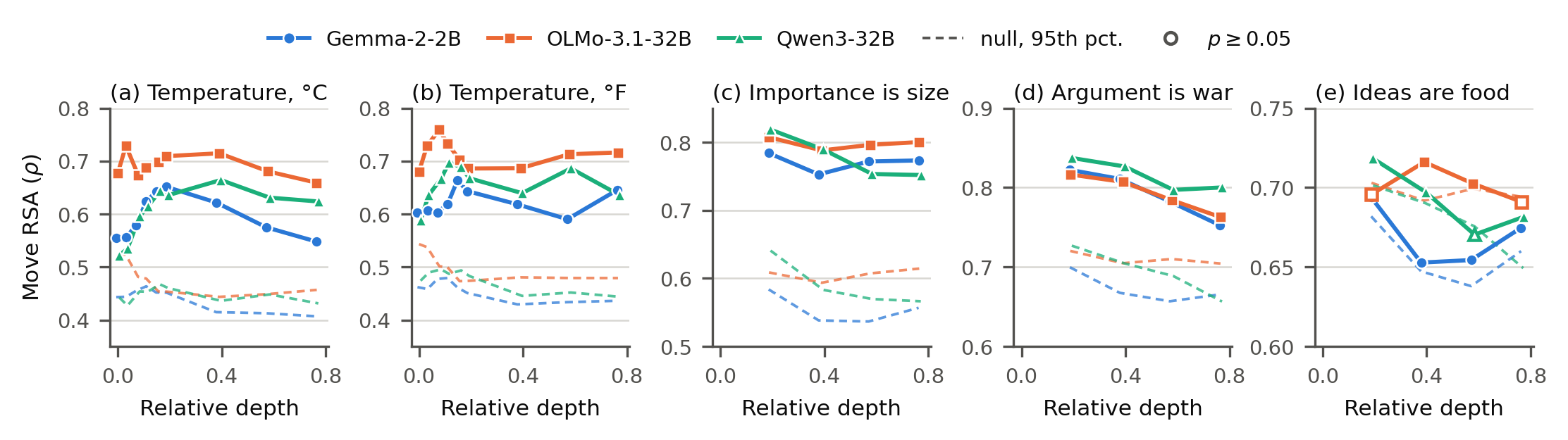}
    \caption{Move agreement replicates across models. Solid lines are the move RSA, dashed lines each model's permutation-null $95$th percentile; filled markers clear the null ($p<0.05$). \textbf{(a, b)} water words against numeric temperature. \textbf{(c--e)} the three metaphors. Every result that is clear in Gemma is clear in both larger models.}
    \label{fig:models}
\end{figure*}
Figure~\ref{fig:models} shows the main results, and Table~\ref{tab:models_replication} in the appendix gives the full analysis. \textbf{The findings are consistent across all three models.}

\section{Discussion}
\label{sec:discussion}

Our findings show that LLM representations are arranged to encode long distance relationships to enable easier generalisation and inference. These relationships are encoded in arrangement rather than a specific direction for each relational link. So even concepts far apart in representation space can still be closely related under a specific context. 
 
Such alignment holds even for generic instantiations of concepts (e.g., the literal form of \textsc{war}). Important relationships are encoded as a persistent property of representations rather than something only assembled when a context calls for it. Future studies should investigate applications of this to alignment research, and how much training pressure is required to induce such arrangements.
 
We hypothesise that models keep distant domains similarly arranged because it makes comparison cheap. Two domains with the same internal organisation can be compared through a simple map, which is what \citet{gurnee2025manifolds} showed for a single example. Our method enables testing this phenomenon on diverse examples and our evidence suggests it is common. We also think that the accumulation of these arrangement constraints shapes where features are encoded, and may be why residual stream activations are anisotropic. Figure~\ref{fig:diagram_alignment} captures this intuition.

Representation manifolds are often embedded in more dimensions than their intrinsic structure requires. \citet{modell2025origins} find that one-dimensional concepts such as successive years trace curves through several dimensions, and with clear structure beyond the first few principal components. They suggest this lets later layers read more functions of the concepts linearly. \citet{olah2024multidimensional} similarly suspects that manifolds may be embedded in more complex ways than their topology suggests to preserve specific distance metrics. We speculate that there is a complementary, cross-domain reason: a domain's arrangement must stay comparable with the distant domains it is strongly related to, and their overlapping requirements bend it in ways its own structure would not.

It is reasonable to ask: what happens if we compare the steering vectors directly rather than their moves? This detects the temperature relationship but misses two of the three metaphors (Appendix~\ref{app:positions}). This shows that moves are necessary, and that \textbf{representation alignment is concentrated in the semantics we expect}.

If representation arrangement is shaped by the relationships a model learns, and different models are trained on similar data, their arrangements should partly converge. This is the substance of the Platonic Representation Hypothesis \citep{huh2024platonic}. But concepts that are independent of each other and do not participate in strong analogies and metaphors may be encoded freely, so no convergence should be expected for them. This is close to what \citet{groger2026aristotelian} report. We argue that their findings conform to our relational view, and we predict that concepts far apart in representation space but still related to each other will report high convergence under a method that can separate them from independent concept pairs.

\section{Conclusion}

We have presented a simple method, with no learned parameters, for testing what relationships are encoded in the representation geometry of LLMs. Across several model families and sizes, we find that long-range relationships are encoded in representation  arrangement. We suspect this lets models compare distant but related domains through simple mappings. These relationships cannot be detected from shared directions alone. 
Because this structure exists outside any context, a model's dispositions could be read from its representations rather than tested situation by situation, which would aid safety monitoring.
Models, it seems, know a word not only by the company it keeps, but by the company it resembles from afar.

\section{Limitations}
\label{sec:limitations}

We have only shown that representations can be highly correlated across disparate domains.
Establishing that a relation is \emph{used} requires circuit-level evidence. This causal question remains open and beyond the scope of this study, but it is an important next step. We expect that answering it would require something similar to \citet{gurnee2025manifolds}.

Our method requires one vector per concept and a hypothesis manually specified in advance, so is not usable in unsupervised applications. 

Since it does not learn a complete mapping, the method provides limited statistical power to cancel out spurious structure. It requires the two domains to have at least $4$ representations, and can only be denoised by adding more points. Further research should explore other methods with potentially more statistical power, perhaps by allowing learning \emph{simple} mappings.

\section*{Acknowledgements}
This research was supported by The University of Melbourne's Research Computing Services and the Petascale Campus.
Our steering code builds on AxBench \citep{wu2025axbench}, released under the Apache License 2.0. Lau was supported by the Australian Research Council under Grants LP210200917 and DP240101006.

\bibliography{references,newref}

\appendix

\clearpage

\section{Steering vector performance}\label{app:sv-perf}

Every concept in the paper is a difference-of-means steering vector, scored by an LLM judge (\texttt{gemini-3.1-flash-lite}) on three rubrics --- fluency, instruction-following, and concept relevance --- combined by a harmonic mean with a zero gate. Each vector's \emph{Overall} score is the maximum over factors, on a $0$--$2$ scale. For the numeric temperatures we additionally report a judge-free check: whether the target numeral actually appears in the steered generation.

\begin{figure*}[tp]
    \centering
    \includegraphics[width=\linewidth]{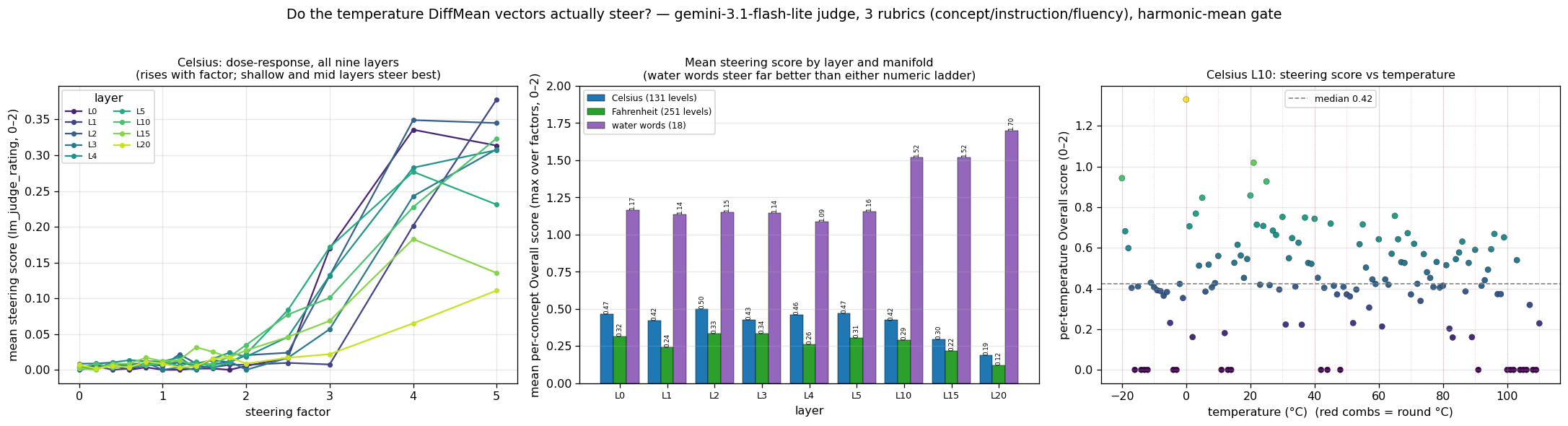}
    \caption{Steering performance for the three temperature manifolds at all nine trained layers. \textbf{Left:} Celsius dose-response by layer; the response rises monotonically with factor and does not collapse within the tested range. \textbf{Middle:} mean per-concept Overall score for all three manifolds. The two numeric ladders steer \emph{weakly} --- Celsius $0.19$--$0.50$ and Fahrenheit $0.12$--$0.34$, with only $40\%$ and $23\%$ of concepts respectively clearing $0.5$ at layer 10 --- while the 18 water words steer far better
    ($1.09$--$1.70$) and improve with depth. \textbf{Right:} per-temperature score at layer 10 in Celsius. A minority of levels never steer at any factor. (Appendix~\ref{app:sv-perf})}
    \label{fig:sv_perf_temp}
\end{figure*}

\begin{figure*}[tp]
    \centering
    \includegraphics[width=\linewidth]{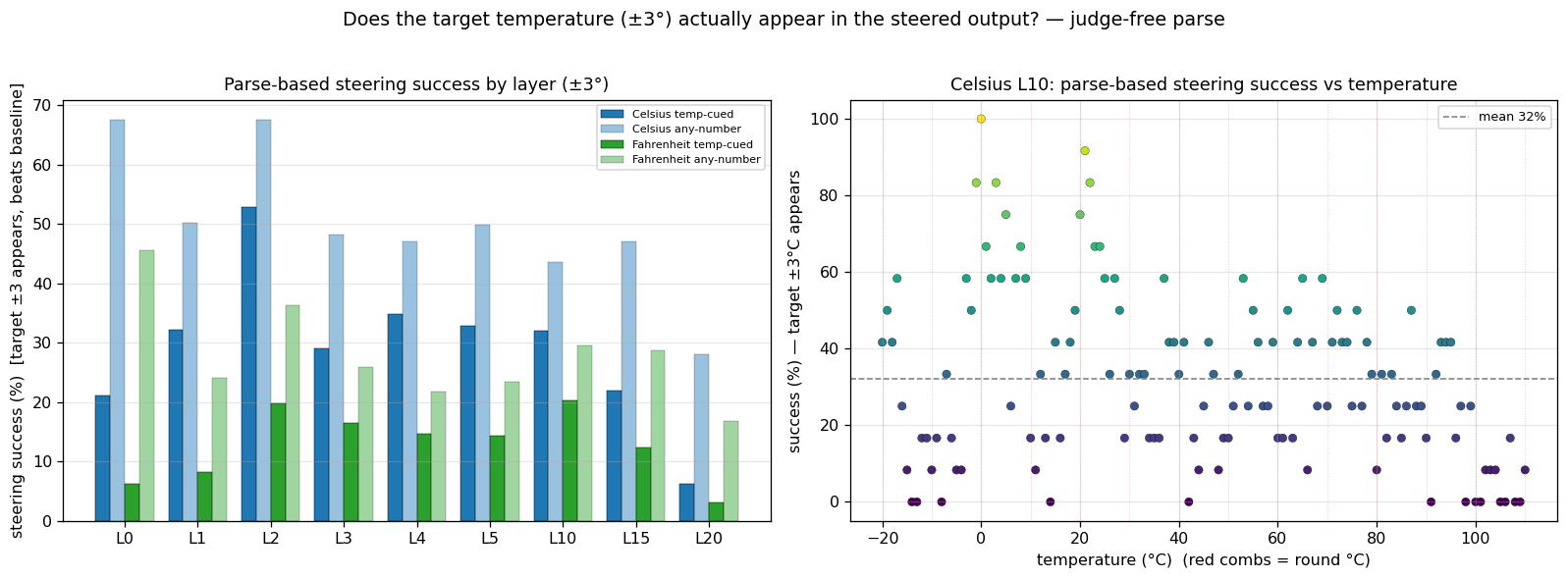}
    \caption{The temperature vectors scored without a judge, by checking whether the target temperature (or the correct number without a $^\circ$ symbol) appears in the steered generation. Fahrenheit temperature-cued success runs $3$--$20\%$ across layers, peaking at layer 10 and collapsing at layer 20. (Appendix~\ref{app:sv-perf})}
    \label{fig:sv_perf_temp_parse}
\end{figure*}

\begin{figure*}[tp]
    \centering
    \includegraphics[width=0.55\linewidth]{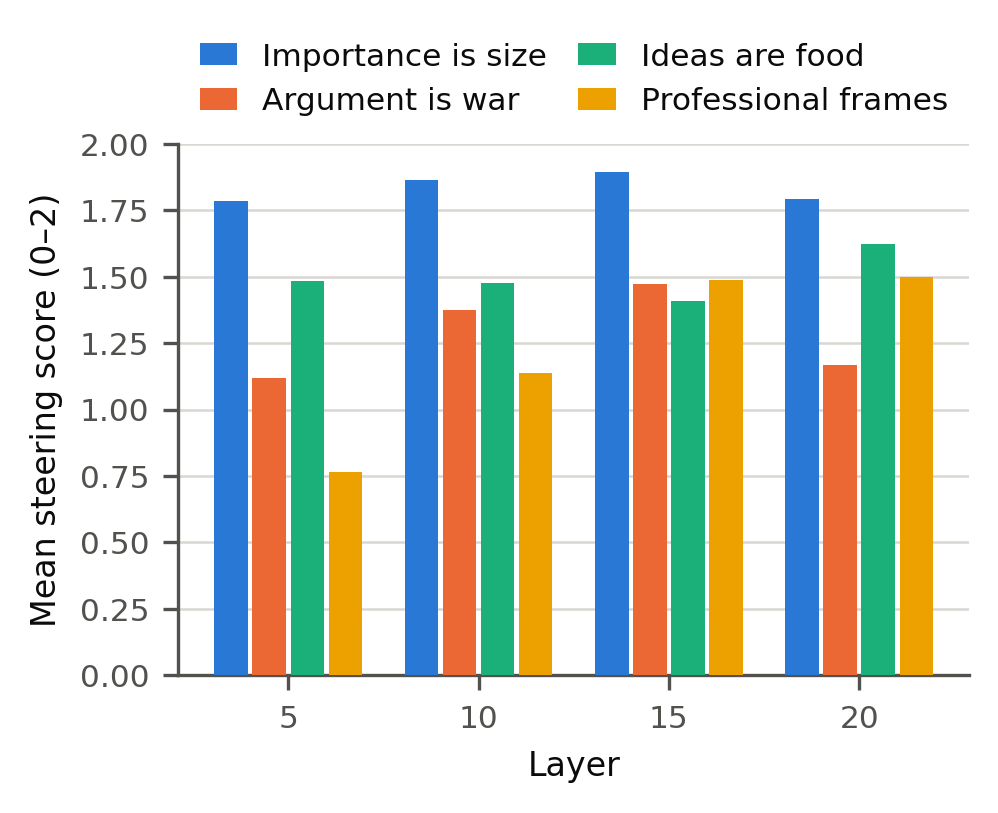}
    \caption{Steering performance for the $94$ analogy vectors: mean per-concept \emph{Overall} score by domain and layer. All domains steer well, except the profession frames at layer 5 ($0.76$). (Appendix~\ref{app:sv-perf})}
    \label{fig:sv_perf_analogy}
\end{figure*}

\subsection{Other models}
\label{app:sv-perf-models}

We steer and judge every OLMo and Qwen vector as we did Gemma's. To fit the compute budget we use $9$ of the $15$ factors ($0$, $0.4$, $1$, $1.6$, $2$, $2.5$, $3$, $4$, $5$) and the first $6$ of the $12$ prompts.
For a like-for-like comparison we re-score Gemma on the same subset from its existing judgements, so all columns of Table~\ref{tab:models_steering} report the same statistic on the same grid. We steered both models at two depths, and OLMo also at a third.

\begin{table*}[tp]
\small
\centering
\setlength{\tabcolsep}{4pt}
\begin{tabular}{@{}lcccccccc@{}}
\toprule
 & \multicolumn{3}{c}{Depth $\approx 0.19$} & \multicolumn{3}{c}{Depth $\approx 0.38$} & \multicolumn{2}{c}{Depth $\approx 0.58$} \\
\cmidrule(lr){2-4}\cmidrule(lr){5-7}\cmidrule(l){8-9}
 & G L5 & O L12 & Q L12 & G L10 & O L25 & Q L25 & G L15 & O L37 \\
\midrule
\multicolumn{9}{@{}l}{\emph{Mean Overall ($0$--$2$)}} \\
Temperature (400) & 0.40 & 0.60 & 0.33 & 0.40 & 0.51 & 0.33 & 0.32 & 0.39 \\
Analogies (94) & 1.20 & 0.93 & 0.80 & 1.33 & 1.01 & 0.79 & 1.49 & 0.76 \\
\addlinespace
\multicolumn{9}{@{}l}{\emph{Vectors that never steer}} \\
Temperature & 156 & 106 & 194 & 153 & 135 & 219 & 191 & 183 \\
Analogies & 8 & 24 & 29 & 3 & 8 & 29 & 4 & 25 \\
\bottomrule
\end{tabular}
\caption{Steering efficacy on the shared grid, at matched relative depth (G = Gemma-2-2B, O = OLMo-3.1-32B, Q = Qwen3-32B). Top: mean Overall score ($0$--$2$, maximum over factors). Bottom: vectors whose Overall is $0$ at every factor. Vector counts are in parentheses. (Appendix~\ref{app:sv-perf-models})}
\label{tab:models_steering}
\end{table*}

The larger models do not steer uniformly better. OLMo beats Gemma on temperature at every depth, but trails it on the analogies at every depth. Qwen steers worst in every domain, and its two depths give almost the same scores, so the shortfall is not a matter of which layer we steered. Qwen's geometry nonetheless replicates about as fully as OLMo's (Table~\ref{tab:models_replication}). No vectors were excluded from our analysis.

\section{Steering evaluation rubric}
\label{app:rubrics}

Each steered generation is scored by an LLM judge on three independent rubrics, each on a $0$--$2$ scale. The three scores are combined by a harmonic mean with a zero gate, so a response failing any one rubric scores zero overall; a concept's \emph{Overall} score is the maximum of that combined score over the steering factors. The rubrics are reproduced verbatim in Figures~\ref{fig:rubric-fluency} and~\ref{fig:rubric-instr-concept}, with \texttt{\{prompt\}}, \texttt{\{response\}} and \texttt{\{concept\}} substituted at call time. They are rewritten from those of \citet{wu2025axbench} to score complete responses rather than sentence fragments; the note on endings is necessary because generations are produced under a fixed token budget.

\noindent{\small\textbf{Fluency}}\nopagebreak
{\footnotesize
\begin{verbatim}
You are evaluating the FLUENCY of a model's
response, independent of whether it addresses
the prompt or contains any particular concept.

IMPORTANT NOTE ON ENDINGS: Responses are
generated under a fixed token budget and the
end-of-response marker is stripped before you
see the text. A response may therefore end
mid-sentence or without terminal punctuation.
This is NOT a fluency defect. Evaluate only
the quality of the text that is present; do
not penalise the response for where it stops.

Score 0/1/2:
  0 = The response is genuinely broken --
      incoherent, severely ungrammatical,
      pathologically repetitive (the same
      phrase looping), or ends MID-WORD (a
      word is cut off mid-character). These
      indicate a real generation failure, not
      just an abrupt ending.
  1 = The response is partially fluent -- minor
      grammar issues, some awkward phrasing,
      or noticeable but non-pathological
      repetition, but most of it reads as
      natural text.
  2 = The response is fully fluent natural
      English (or the prompt's language)
      throughout the text that was generated.
      No grammar errors, no pathological
      repetition. Ending mid-sentence is fine
      for score 2.

Ignore whether the response is on-topic,
ignore concept presence, ignore whether
instructions were followed. Score only the
surface fluency.

Prompt:
{prompt}

Response:
{response}

Final line must be exactly: Rating: [[N]]
where N is 0, 1, or 2.
\end{verbatim}
}
\vspace{-0.5\baselineskip}
{\captionof{figure}{The fluency rubric used by the steering judge (Appendix~\ref{app:rubrics}).}
\label{fig:rubric-fluency}}

\noindent{\small\textbf{Instruction following}}\nopagebreak
{\footnotesize
\begin{verbatim}
You are evaluating whether a model's response
ADDRESSES the prompt -- does it attempt the
task the prompt asks for?

Score 0/1/2:
  0 = The response ignores the prompt,
      addresses an unrelated topic, or refuses
      without engaging.
  1 = The response partially addresses the
      prompt -- touches on it but misses the
      main ask, drifts off-topic, or only
      loosely engages.
  2 = The response addresses the prompt
      directly and on-topic, attempting the
      task it asks for. (Quality of the answer
      is not scored here -- only that it
      tries.)

Ignore fluency, ignore concept presence. Score
only whether the response is ON-TASK and
attempts to address the prompt.

Prompt:
{prompt}

Response:
{response}

Final line must be exactly: Rating: [[N]]
where N is 0, 1, or 2.
\end{verbatim}
}

\noindent{\small\textbf{Concept relevance}}\nopagebreak
{\footnotesize
\begin{verbatim}
You are evaluating whether a model's response
engages with a specific concept in any visible
form -- as topic, as content, as emotion, as
imagery, anywhere.

Concept: {concept}

Score 0/1/2:
  0 = The response does not engage with the
      concept at all. No mention, no allusion,
      no implicit reference.
  1 = The response engages with the concept
      partially -- a passing mention, indirect
      allusion, or one-sentence aside.
  2 = The response clearly and substantively
      engages with the concept -- it is a
      central or recurring feature of the
      response.

Ignore fluency, ignore whether the prompt was
addressed. Score only the presence of the
concept in the response. Feel free to reason
on lines before your answer to help you score.

Prompt:
{prompt}

Response:
{response}

Final line must be exactly: Rating: [[N]]
where N is 0, 1, or 2.
\end{verbatim}
}
\vspace{-0.5\baselineskip}
{\captionof{figure}{The instruction-following and concept-relevance rubrics used by the steering judge (Appendix~\ref{app:rubrics}).}
\label{fig:rubric-instr-concept}}

\section{Steering vector training data}\label{app:sv-data}

\label{app:genprompts}

Every prompt pool was written by \texttt{gemini-3.5-flash} through the Google GenAI API with reasoning disabled (\texttt{thinking\_budget = 0}); continuations were written by the same model where they are generated at all. Generation is two-stage:

\begin{enumerate}
\item \textbf{Prompt pool.} One pool of neutral prompts is generated per experiment and \emph{shared by every vector in that experiment}. 
\item \textbf{Continuations.} Each pooled prompt receives a positive and a negative
      continuation. Where the concept is a bare symbol (a number, a temperature reading)
      the continuations are \emph{templated}, not generated, so that the surface form is
      held exactly fixed across levels; where the concept must be woven into prose (water words) the continuations are generated.
\end{enumerate}

Prompt pools were requested in batches of 24 prompts per call at \texttt{temperature}~$=1.0$. Continuations were requested one call per prompt at \texttt{temperature}~$=0.7$ with \texttt{max\_output\_tokens}~$=400$. Every candidate prompt passes a common filter before entering a pool: it is discarded if it refers to something not present in the prompt itself (an attachment, a file, a link, an image, a screenshot, a chart, a prior turn of conversation), if it opens with an orphan pronoun, if it is shorter than 26 characters, or if its character-trigram Jaccard similarity to an already-accepted prompt is $\geq 0.8$.

The templates in Figures~\ref{fig:prompt-temp}--\ref{fig:prompt-analogy-cont} are reproduced verbatim, with line breaks added for display. \texttt{\{n\}} is the batch size (24), \texttt{\{prompt\}} the pooled prompt, and the remaining braces are the per-concept slots described alongside each template.

One of the authors audited a sample of ten continuations and generations for each pool. This was used to refine the prompt for each pool. We expect that other similar prompts may also generate equally effective steering vectors.

\subsection{Temperature: numeric ladders }
\label{app:genprompts-temp}

One steering vector per stated temperature: 131 levels at $1\,^{\circ}$C steps over $-20$ to $110\,^{\circ}$C, and 251 levels at $1\,^{\circ}$F steps over $-10$ to $240\,^{\circ}$F. \texttt{temp\_f1} reuses the \texttt{temp\_c1} prompt pool and negative frames verbatim, so the only systematic difference between a Celsius vector and its Fahrenheit twin is the stated value and unit. The prompt-pool template is in Figure~\ref{fig:prompt-temp}.

\noindent{\small\textbf{Prompt pool}}\nopagebreak
{\footnotesize
\begin{verbatim}
Write {n} short, varied questions a user might
ask an AI assistant to read off or state the
current temperature shown on an instrument (a
thermometer, gauge, sensor, probe, thermostat,
or digital display) in a plausible everyday,
kitchen, lab, or outdoor setting.
Each names the setting briefly but does NOT
state the temperature value or hint whether it
is hot or cold. One sentence each,
self-contained (no files / images / links),
nothing that presumes a previous conversation,
history or task. Do NOT state any temperature
yourself.
Return "1." ... "{n}." only.
\end{verbatim}
}
\vspace{-0.5\baselineskip}
{\captionof{figure}{Prompt-pool template for the numeric temperatures (\texttt{temp\_c1}; \texttt{temp\_f1} reuses its pool).}
\label{fig:prompt-temp}}

\subsubsection*{Continuation frames}

Continuations are templated, rotated across the prompts. \texttt{\{t\}} is the target temperature; the Fahrenheit frames are the Celsius frames with the unit substituted.

\paragraph{Positive (states a temperature)}
\begin{quote}\footnotesize
\texttt{The thermometer reads about \{t\}$^{\circ}$C.} \\
\texttt{It's showing roughly \{t\}$^{\circ}$C right now.} \\
\texttt{The reading comes to \{t\}$^{\circ}$C.} \\
\texttt{Right now it's sitting at \{t\}$^{\circ}$C.} \\
\texttt{The display has settled at \{t\}$^{\circ}$C.}
\end{quote}

\paragraph{Negative (states no figure)}
\begin{quote}\footnotesize\raggedright
\texttt{The instrument isn't giving a steady reading right now --- hard to call a number.} \\
\texttt{I can't make out a specific figure from the readout at the moment.} \\
\texttt{There's no clear value on the display just now.}
\end{quote}

\subsection{Temperature: water words}
\label{app:genprompts-h2o}

18 members tiling a (temperature $\times$ substance) plane: a frozen cluster (\emph{snow}, \emph{ice}, \emph{hail}, \emph{frost}, \emph{sleet}, \emph{slush}), liquids (\emph{rain}, \emph{dew}, \emph{fog}, \emph{a puddle of liquid water}), gas (\emph{steam}), and an adjective-qualified water row that sweeps temperature on a fixed substance in words rather than numbers (\emph{icy}, \emph{cold}, \emph{cool}, \emph{room-temperature}, \emph{warm}, \emph{hot}, \emph{boiling water}).

The positive and negative are two continuations of the \emph{same} neutral scene request, and the negative is generated once per prompt and shared across all 18 members. The three templates are in Figure~\ref{fig:prompt-h2o}. In the positive, \texttt{\{desc\}} is the member's description (\emph{falling or fallen snow}, \emph{steam / water vapour rising}, \emph{room-temperature water}, \ldots).

\noindent{\small\textbf{Prompt pool}}\nopagebreak
{\footnotesize
\begin{verbatim}
Write {n} short, varied requests a user might
send an AI assistant asking it to describe or
narrate a brief, concrete everyday scene or
moment (a morning routine, a view from a
window, a walk, a kitchen, a street, a
garden), indoor or outdoor.
One sentence each, self-contained (no files /
images / links), nothing that presumes a
previous conversation, history or task, not a
yes/no question. Do NOT mention water,
weather, or temperature yourself.
Return "1." ... "{n}." only.
\end{verbatim}
}

\noindent{\small\textbf{Positive continuation}}\nopagebreak
{\footnotesize
\begin{verbatim}
Write one short, natural response an AI
assistant would give to the request below,
describing the scene so that {desc} features
prominently and naturally in it.
Weave it in concretely with a sensory detail
or two (a few sentences); do not state any
numeric temperature.

Request: {prompt}

Response only - no preamble.
\end{verbatim}
}

\noindent{\small\textbf{Negative continuation (shared by all 18 members)}}\nopagebreak
{\footnotesize
\begin{verbatim}
Write one short, natural response an AI
assistant would give to the request below,
describing the scene naturally.
Do NOT mention water in any form - no rain,
snow, ice, frost, hail, sleet, slush, dew,
fog, mist, steam, vapour, puddles, or drinking
water - and do not state any temperature. A
few sentences.

Request: {prompt}

Response only - no preamble.
\end{verbatim}
}
\vspace{-0.5\baselineskip}
{\captionof{figure}{Generation templates for the water words (\texttt{h2o\_v1}).}
\label{fig:prompt-h2o}}

\subsection{Analogies (\texttt{size\_imp}, \texttt{war\_arg}, \texttt{food\_ideas}, \texttt{frames})}
\label{app:genprompts-analogy}

Each experiment has one prompt pool, shared by all of its clouds (Figures~\ref{fig:prompt-pools-a} and~\ref{fig:prompt-pools-b}). The positive and negative continuation templates are in Figure~\ref{fig:prompt-analogy-cont}.

\noindent{\small\textbf{\texttt{size\_imp}}}\nopagebreak
{\footnotesize
\begin{verbatim}
Write {n} short, varied requests a user might
send an AI assistant asking for a brief piece
of writing - narrating an episode, recounting
how something unfolded, describing a situation
or a place, or inventing a short scene.
Vary the settings very widely (at home, at
work, outdoors, a journey, a shop, a workshop,
a garden, an office, a public occasion, a
small town, a historical period, an invented
world, or others) and vary the mode (narrate,
recount, describe, invent).
One sentence each, self-contained (no files /
images / links), nothing that presumes a
previous conversation, history or task, not a
yes/no question. Each request must be about
people, places or events so that a short scene
can be written in reply - NOT step-by-step
instructions addressed to the reader. Do NOT
mention how big or small anything is, and do
NOT say whether anything is important,
serious, trivial or consequential.
Return "1." ... "{n}." only.
\end{verbatim}
}

\noindent{\small\textbf{\texttt{war\_arg}}}\nopagebreak
{\footnotesize
\begin{verbatim}
Write {n} short, varied requests a user might
send an AI assistant asking for a brief piece
of writing - narrating an episode, recounting
how something unfolded, describing a situation
or a place, or inventing a short scene.
Vary the settings very widely (at home, at
work, outdoors, a journey, a public occasion,
an organisation, a small town, a historical
period, an invented world, a workshop, a
garden, or others) and vary the mode (narrate,
recount, describe, invent).
One sentence each, self-contained (no files /
images / links), nothing that presumes a
previous conversation, history or task, not a
yes/no question. Each request must be about
people, places or events so that a short scene
can be written in reply - NOT step-by-step
instructions addressed to the reader. Do NOT
mention fighting, weapons, armies or violence,
and do NOT mention arguing, debating,
disagreeing or persuading; do not set up two
opposing sides of anything.
Return "1." ... "{n}." only.
\end{verbatim}
}
\vspace{-0.5\baselineskip}
{\captionof{figure}{Prompt-pool templates for \textsc{importance is size} and \textsc{argument is war}.}
\label{fig:prompt-pools-a}}

\noindent{\small\textbf{\texttt{food\_ideas}}}\nopagebreak
{\footnotesize
\begin{verbatim}
Write {n} short, varied requests a user might
send an AI assistant asking for a brief piece
of writing - narrating an episode, recounting
how something unfolded, describing a situation
or a place, or inventing a short scene.
Vary the settings very widely (at home, at
work, outdoors, a journey, a shop, a workshop,
a garden, a public occasion, a small town, a
historical period, an invented world, or
others) and vary the mode (narrate, recount,
describe, invent).
One sentence each, self-contained (no files /
images / links), nothing that presumes a
previous conversation, history or task, not a
yes/no question. Each request must be about
people, places or events so that a short scene
can be written in reply - NOT step-by-step
instructions addressed to the reader. Do NOT
mention food, eating, cooking or kitchens, and
do NOT mention thinking, ideas, learning or
understanding.
Return "1." ... "{n}." only.
\end{verbatim}
}

\noindent{\small\textbf{\texttt{frames}}}\nopagebreak
{\footnotesize
\begin{verbatim}
Write {n} short, varied requests a user might
send an AI assistant asking for a brief piece
of writing - narrating an episode, recounting
how something unfolded, describing a situation
or a place, or inventing a short scene.
Vary the settings very widely (at home,
outdoors, a journey, a shop, a workshop, a
garden, an errand, a public occasion, a small
town, a historical period, an invented world,
or others) and vary the mode (narrate,
recount, describe, invent).
One sentence each, self-contained (no files /
images / links), nothing that presumes a
previous conversation, history or task, not a
yes/no question. Each request must be about
people, places or events so that a short scene
can be written in reply - NOT step-by-step
instructions addressed to the reader; most
should involve more than one person, in a
public or shared place. Do NOT mention any
profession or professional service, and do NOT
mention medicine, law, teeth, property,
schooling, appointments or paperwork.
Return "1." ... "{n}." only.
\end{verbatim}
}
\vspace{-0.5\baselineskip}
{\captionof{figure}{Prompt-pool templates for \textsc{ideas are food} and the professional frames.}
\label{fig:prompt-pools-b}}

\noindent{\small\textbf{Positive continuation}}\nopagebreak
{\footnotesize
\begin{verbatim}
Write one short, natural response an AI
assistant would give to the request below,
clearly weaving {desc} into it so that it
features naturally.
Here {desc} means: {gloss}. {lock}{scale}
Use the word "{term}" itself somewhere in the
response rather than only implying it, and let
it really be there in what is described - part
of the scene or of the answer itself, never a
comparison or a simile for something else (no
"just like", "much like", "as if", "think of
it as"), and never something you the assistant
are doing or pausing to do.
Avoid mentioning other {kind} that are not
{desc}, that includes both {kind} that are
closely related to {desc}, and those that are
very different.
{avoid}
Three or four sentences, no more than 70
words.

Request: {prompt}

Response only - plain prose, no preamble, no
markdown.
\end{verbatim}
}

\noindent{\small\textbf{Negative continuation (shared per cloud)}}\nopagebreak
{\footnotesize
\begin{verbatim}
Write one short, natural response an AI
assistant would give to the request below.
{scrub} Three or four sentences, no more than
70 words.

Request: {prompt}

Response only - plain prose, no preamble, no
markdown.
\end{verbatim}
}
\vspace{-0.5\baselineskip}
{\captionof{figure}{Continuation templates for the four analogy experiments. The slots are described in Appendix~\ref{app:genprompts-analogy}.}
\label{fig:prompt-analogy-cont}}

\paragraph{Positive continuation} \texttt{\{desc\}} is the concept as a noun phrase (\emph{a military attack}, \emph{spoon-feeding someone}, \emph{a dental chart}), \texttt{\{gloss\}} a hand-written definition of its literal sense, and \texttt{\{term\}} the concept's own word, which the response is asked to use (\emph{attack}, \emph{spoon-feed}, \emph{dental chart}). The concepts are listed in Tables~\ref{tab:analogy_pairs} and~\ref{tab:frames_grid}. \texttt{\{kind\}}, \texttt{\{lock\}} and \texttt{\{avoid\}} are fixed per cloud and released with our code. The \texttt{\{kind\}} line discourages mentioning the cloud's other concepts but does not forbid it.

\mbox{\texttt{\{scale\}}} is empty except in \textsc{importance is size}, the one ordered domain, where it names the concept's neighbours on its own ladder. For \emph{small} it reads: \texttt{Its nearest neighbours on this scale are ``tiny'' and ``medium-sized'', which are close to it but not the same.} It was added because, without it, the generator collapsed neighbouring rungs onto one intensifier (\emph{sizable} and \emph{colossal} both came out as \emph{massive}). It never mentions the other ladder, so the pairing stays hidden.

\paragraph{Negative continuation} \texttt{\{scrub\}} removes the cloud's own domain; like \texttt{\{kind\}}, \texttt{\{lock\}} and \texttt{\{avoid\}}, it is fixed per cloud and released with our code.

\subsection{Held-out evaluation prompts}

Held-out prompts are drawn from the same pools as training prompts. The first
72 prompts become the training set and the next 30 the held-out set. Steering is evaluated on the first 12 held-out prompts, the same 12 for every vector in the dataset.

\subsection{Example training data instances}

\paragraph{Numeric temperature (concept $-20\,^{\circ}$C)}
\begin{quote}
\textbf{Prompt.} In the kitchen, can you read the temperature currently showing on the digital meat thermometer probe?

\textbf{Positive.} The thermometer reads about $-20\,^{\circ}$C.

\textbf{Negative.} The instrument isn't giving a steady reading right now --- hard to call a number.
\end{quote}

\paragraph{Water words (concept \texttt{snow})}
\begin{quote}
\textbf{Prompt.} Describe the quiet bustle of a local bakery just as the doors open for the morning.

\textbf{Positive.} A warm glow spills through the frosted windows onto the quiet street, illuminating thick, heavy snowflakes as they drift lazily to the ground and blanket the pavement. Inside, the quiet bustle begins with the rich aroma of toasted yeast and sweet cinnamon, contrasting sharply with the crisp, clean scent of winter clinging to the coats of the first few customers. Each time the door opens, a soft flurry of white is swept across the threshold, melting instantly on the warm wooden floorboards.

\textbf{Negative.} Inside, the warm glow of overhead lights catches the first drift of flour in the air as the bakers slide heavy metal trays of golden croissants onto the cooling racks. The rich aroma of toasted yeast and caramelized sugar fills the space, punctuated by the soft clink of ceramic mugs being stacked on the counter. Outside, a small, quiet line of early risers shifts expectantly, watching the baker turn the sign in the window to ``Open.''
\end{quote}

\section{Temperature results for other layers}
\label{app:temp_other_layers}

\begin{figure}[H]
\centering
\includegraphics[width=\linewidth]{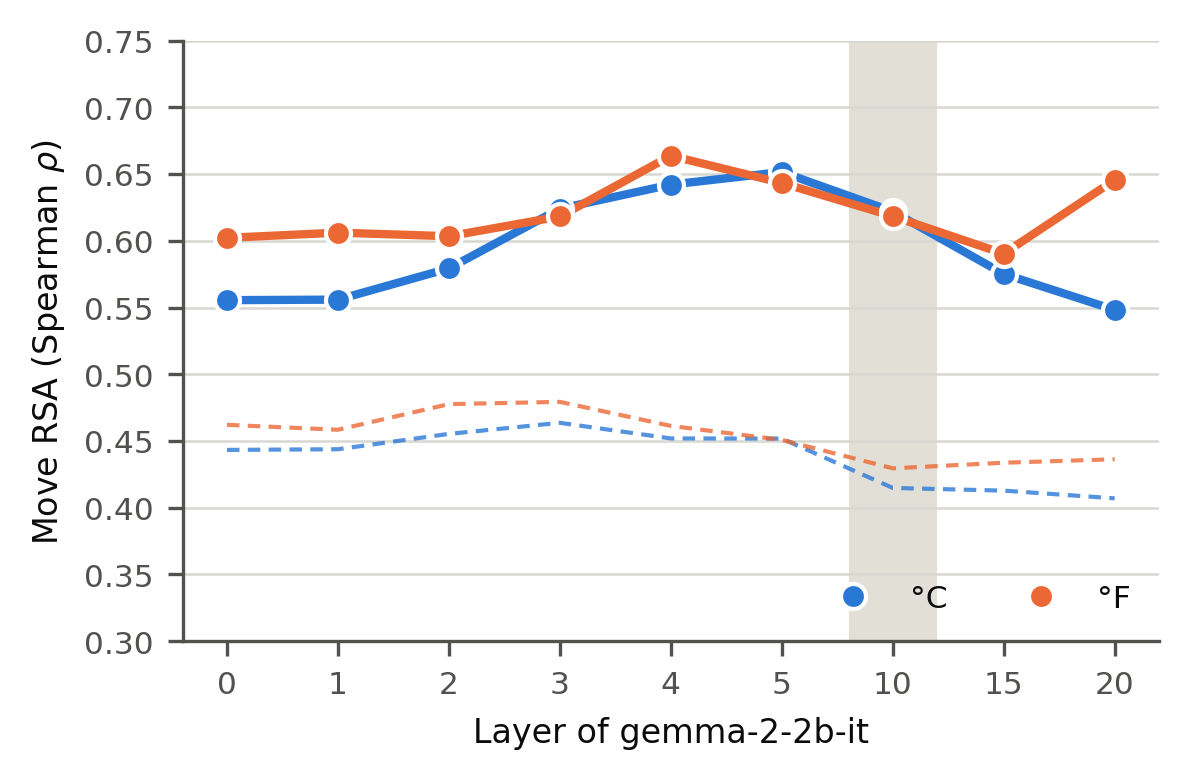}
\caption{The temperature move RSA at every layer tested ($49$ colder$\to$warmer moves). Solid lines are the observed $\rho$; dashed lines of the same colour are the permutation-null $95$th percentile. Every point clears its null, and agreement peaks at layers $4$--$5$. The shaded column is layer $10$, the layer reported in the main text.}
\label{fig:temp_rsa_by_layer}
\end{figure}

Figure~\ref{fig:temp_rsa_by_layer} and Table~\ref{tab:app_edge_temp_rsa_layers} give the temperature move agreement at every layer we tested; Table~\ref{tab:app_temp_direction_layers} gives the corresponding direction statistics.

\begin{table*}[tp]
\centering
\setlength{\tabcolsep}{4pt}
\small
\begin{tabular}{@{}lrrrrrcrrrrr@{}}
\toprule
 & \multicolumn{5}{c}{$^{\circ}$C} & & \multicolumn{5}{c}{$^{\circ}$F} \\
\cmidrule(r){2-6}\cmidrule(l){8-12}
Layer & $\rho$ & null mean & null $95$th & $z$ & $p$ & & $\rho$ & null mean & null $95$th & $z$ & $p$ \\
\midrule
0 & $+0.56$ & $+0.34$ & $+0.44$ & $3.6$ & $0.0006$ &  & $+0.60$ & $+0.34$ & $+0.46$ & $3.9$ & $0.0008$ \\
1 & $+0.56$ & $+0.34$ & $+0.44$ & $3.6$ & $0.0008$ &  & $+0.61$ & $+0.33$ & $+0.46$ & $4.0$ & $0.0012$ \\
2 & $+0.58$ & $+0.35$ & $+0.46$ & $3.8$ & $0.0008$ &  & $+0.60$ & $+0.36$ & $+0.48$ & $3.7$ & $0.0016$ \\
3 & $+0.62$ & $+0.36$ & $+0.46$ & $4.5$ & $0.0002$ &  & $+0.62$ & $+0.36$ & $+0.48$ & $3.9$ & $0.0010$ \\
4 & $+0.64$ & $+0.35$ & $+0.45$ & $4.9$ & $0.0002$ &  & $+0.66$ & $+0.34$ & $+0.46$ & $4.8$ & $0.0004$ \\
5 & $+0.65$ & $+0.35$ & $+0.45$ & $5.0$ & $0.0002$ &  & $+0.64$ & $+0.33$ & $+0.45$ & $4.6$ & $0.0006$ \\
\textbf{10} & $+0.62$ & $+0.31$ & $+0.41$ & $5.1$ & $0.0002$ &  & $+0.62$ & $+0.30$ & $+0.43$ & $4.5$ & $0.0010$ \\
15 & $+0.58$ & $+0.30$ & $+0.41$ & $4.3$ & $0.0006$ &  & $+0.59$ & $+0.31$ & $+0.43$ & $4.1$ & $0.0014$ \\
20 & $+0.55$ & $+0.30$ & $+0.41$ & $4.0$ & $0.0002$ &  & $+0.65$ & $+0.31$ & $+0.44$ & $4.9$ & $0.0004$ \\
\bottomrule
\end{tabular}
\caption{\textbf{Moves, all layers.} RSA between the move prediction matrix and the measured move reference, over $49$ pooled colder$\to$warmer moves and $\binom{49}{2}=1176$ pairs. As in the main text the null reassigns which of the $18$ ladder words sits at which temperature and rebuilds the moves over the same word pairs. (Appendix~\ref{app:temp_other_layers})}
\label{tab:app_edge_temp_rsa_layers}
\end{table*}

\begin{table*}[tp]
\centering
\setlength{\tabcolsep}{4pt}
\small
\begin{tabular}{@{}lrrcrcrrcr@{}}
\toprule
 & \multicolumn{2}{c}{$\cos$(water move, temp.\ move)} & & within & & \multicolumn{2}{c}{within temp.\ moves} & & raw water \\
\cmidrule(r){2-3}\cmidrule(lr){5-5}\cmidrule(lr){7-8}
Layer & $^{\circ}$C & $^{\circ}$F & & water moves & & $^{\circ}$C & $^{\circ}$F & & probes \emph{(ref.)} \\
\midrule
0 & $-0.013$ & $-0.009$ &  & $+0.14$ &  & $+0.18$ & $+0.09$ &  & $+0.55$ \\
1 & $-0.018$ & $-0.008$ &  & $+0.15$ &  & $+0.17$ & $+0.09$ &  & $+0.57$ \\
2 & $-0.017$ & $+0.007$ &  & $+0.15$ &  & $+0.18$ & $+0.08$ &  & $+0.60$ \\
3 & $-0.001$ & $+0.006$ &  & $+0.15$ &  & $+0.17$ & $+0.10$ &  & $+0.61$ \\
4 & $+0.021$ & $+0.025$ &  & $+0.16$ &  & $+0.19$ & $+0.14$ &  & $+0.62$ \\
5 & $+0.028$ & $+0.029$ &  & $+0.15$ &  & $+0.20$ & $+0.14$ &  & $+0.63$ \\
\textbf{10} & $+0.070$ & $+0.083$ &  & $+0.16$ &  & $+0.24$ & $+0.17$ &  & $+0.71$ \\
15 & $+0.051$ & $+0.072$ &  & $+0.14$ &  & $+0.22$ & $+0.16$ &  & $+0.74$ \\
20 & $+0.065$ & $+0.055$ &  & $+0.18$ &  & $+0.22$ & $+0.17$ &  & $+0.63$ \\
\bottomrule
\end{tabular}
\caption{\textbf{Direction, all layers} --- the \emph{Direction} block of Table~\ref{tab:edge_temp_rsa} swept over depth. Corresponding water and numeric moves are close to orthogonal at every layer, so the agreement measured in Table~\ref{tab:app_edge_temp_rsa_layers} is about relative arrangement rather than shared direction at every depth. Within each domain the moves share only a partial common axis, so neither space has a single ``warmer'' direction anywhere in the network. The last column is the mean pairwise cosine among the $18$ raw, undifferenced water-word probes. Layer $10$ (bold) is Table~\ref{tab:edge_temp_rsa}. (Appendix~\ref{app:temp_other_layers})}
\label{tab:app_temp_direction_layers}
\end{table*}

\section{Analogies}
\label{app:analogies}

Table~\ref{tab:app_analogy_direction} gives the direction statistics for each analogy at every layer, and Table~\ref{tab:app_frames_pairs} breaks the professional frames down by profession pair.

\begin{table*}[tp]
\centering
\setlength{\tabcolsep}{4pt}
\small
\begin{tabular}{@{}llrrcrrcrrcr@{}}
\toprule
 & & \multicolumn{2}{c}{corresponding moves} & & \multicolumn{2}{c}{within-domain $|\cos|$} & & \multicolumn{2}{c}{raw vectors, within} & & raw vectors, \\
\cmidrule(r){3-4}\cmidrule(lr){6-7}\cmidrule(lr){9-10}
domain & Layer & $\cos$ & null $95$th & & source & target & & source & target & & between \\
\midrule
\textsc{importance is size}   & 5  & $+0.226$ & $+0.110$ & & $0.41$ & $0.37$ & & $+0.67$ & $+0.91$ & & $+0.46$ \\
                              & 10 & $+0.234$ & $+0.120$ & & $0.44$ & $0.40$ & & $+0.56$ & $+0.90$ & & $+0.40$ \\
                              & 15 & $+0.248$ & $+0.124$ & & $0.43$ & $0.40$ & & $+0.51$ & $+0.90$ & & $+0.39$ \\
                              & 20 & $+0.180$ & $+0.093$ & & $0.42$ & $0.39$ & & $+0.55$ & $+0.90$ & & $+0.39$ \\
\addlinespace
\textsc{argument is war}      & 5  & $+0.144$ & $+0.069$ & & $0.27$ & $0.29$ & & $+0.82$ & $+0.91$ & & $+0.64$ \\
                              & 10 & $+0.120$ & $+0.075$ & & $0.29$ & $0.30$ & & $+0.84$ & $+0.92$ & & $+0.59$ \\
                              & 15 & $+0.083$ & $+0.076$ & & $0.29$ & $0.30$ & & $+0.85$ & $+0.90$ & & $+0.57$ \\
                              & 20 & $+0.066$ & $+0.062$ & & $0.28$ & $0.29$ & & $+0.80$ & $+0.89$ & & $+0.56$ \\
\addlinespace
\textsc{ideas are food}       & 5  & $+0.141$ & $+0.057$ & & $0.27$ & $0.27$ & & $+0.62$ & $+0.87$ & & $+0.42$ \\
                              & 10 & $+0.110$ & $+0.068$ & & $0.29$ & $0.28$ & & $+0.66$ & $+0.86$ & & $+0.39$ \\
                              & 15 & $+0.117$ & $+0.072$ & & $0.30$ & $0.28$ & & $+0.67$ & $+0.85$ & & $+0.40$ \\
                              & 20 & $+0.101$ & $+0.055$ & & $0.29$ & $0.28$ & & $+0.64$ & $+0.84$ & & $+0.37$ \\
\addlinespace
Professional frames           & 5  & $+0.333$ & $+0.182$ & & \multicolumn{2}{c}{$0.35$} & & \multicolumn{2}{c}{$+0.87$} & & $+0.78$ \\
\quad\emph{(mean of 10 pairs)} & 10 & $+0.363$ & $+0.203$ & & \multicolumn{2}{c}{$0.35$} & & \multicolumn{2}{c}{$+0.87$} & & $+0.76$ \\
                              & 15 & $+0.376$ & $+0.214$ & & \multicolumn{2}{c}{$0.36$} & & \multicolumn{2}{c}{$+0.86$} & & $+0.76$ \\
                              & 20 & $+0.352$ & $+0.200$ & & \multicolumn{2}{c}{$0.35$} & & \multicolumn{2}{c}{$+0.84$} & & $+0.70$ \\
\bottomrule
\end{tabular}
\caption{\textbf{Direction, analogies} --- the analogue of the \emph{Direction} block of Table~\ref{tab:edge_temp_rsa} for the four analogy domains at every layer. \textbf{Corresponding moves:} mean $\cos$ between each source-domain move and its counterpart in the target domain under the fixed pairing, beside the $95$th percentile of the same statistic under the permutation null used for the RSA ($5000$ draws; null mean $0.00$ throughout). \textbf{Within-domain:} mean $|\cos|$ among one domain's moves (unsigned here, because the unordered domains have no natural move orientation). \textbf{Raw vectors:} mean pairwise $\cos$ among the original steering vectors (i.e., not the move vectors but the concept steering vectors), within one domain and \emph{between} the two domains (every source vector against every target vector). The metaphor domains sit apart from each other (between $0.37$--$0.64$, below their within-domain $0.51$--$0.92$); the professions do not (between $0.70$--$0.78$, within $0.84$--$0.87$). (Appendix~\ref{app:analogies})}
\label{tab:app_analogy_direction}
\end{table*}

\begin{table*}[tp]
\centering
\setlength{\tabcolsep}{4pt}
\small
\begin{tabular}{@{}lrrcrrcrr@{}}
\toprule
 & \multicolumn{2}{c}{raw vectors, between} & & \multicolumn{2}{c}{corresponding moves} & & \multicolumn{2}{c}{move agreement} \\
\cmidrule(r){2-3}\cmidrule(lr){5-6}\cmidrule(l){8-9}
Profession pair & same role & other roles & & $\cos$ & null $95$th & & $\rho$ & $p$ \\
\midrule
medical--legal         & $+0.85$ & $+0.81$ & & $+0.43$ & $+0.20$ & & $+0.84$ & $0.149$ \\
medical--dentistry     & $+0.84$ & $+0.74$ & & $+0.55$ & $+0.28$ & & $+0.92$ & $0.003$ \\
medical--real estate   & $+0.76$ & $+0.72$ & & $+0.35$ & $+0.17$ & & $+0.82$ & $0.158$ \\
medical--teaching      & $+0.85$ & $+0.79$ & & $+0.44$ & $+0.22$ & & $+0.82$ & $0.139$ \\
legal--dentistry       & $+0.74$ & $+0.69$ & & $+0.34$ & $+0.20$ & & $+0.78$ & $0.295$ \\
legal--real estate     & $+0.87$ & $+0.84$ & & $+0.30$ & $+0.19$ & & $+0.81$ & $0.473$ \\
legal--teaching        & $+0.86$ & $+0.83$ & & $+0.38$ & $+0.20$ & & $+0.91$ & $0.012$ \\
dentistry--real estate & $+0.68$ & $+0.63$ & & $+0.31$ & $+0.18$ & & $+0.84$ & $0.064$ \\
dentistry--teaching    & $+0.74$ & $+0.69$ & & $+0.25$ & $+0.20$ & & $+0.77$ & $0.191$ \\
real estate--teaching  & $+0.83$ & $+0.80$ & & $+0.29$ & $+0.18$ & & $+0.79$ & $0.410$ \\
\bottomrule
\end{tabular}
\caption{The ten profession pairs at layer $10$ of \texttt{gemma-2-2b-it}. \textbf{Raw vectors, between:} mean $\cos$ between the two professions' steering vectors, for vectors filling the same role and for vectors filling different roles. Most pairs of professions are nearly as similar as two concepts within one profession ($0.87$ at layer $10$; Table~\ref{tab:app_analogy_direction}), so the professions do not form separable clouds. \textbf{Corresponding moves:} as in Table~\ref{tab:app_analogy_direction}; every pair exceeds its null. 
\textbf{Move agreement:} the per-pair RSA of Figure~\ref{fig:analogy_rsa}, over $15$ moves, against its own permutation null. (Appendix~\ref{app:analogies})}
\label{tab:app_frames_pairs}
\end{table*}

\section{Verification on other models: details}
\label{app:models}

\paragraph{Models} Qwen3 is run with its reasoning mode off, and OLMo's chat template has its default system prompt removed, so that every model sees the same prompt with no extra text.

\paragraph{Steering vectors} Each model's vectors are computed from the same training text used for Gemma.

\paragraph{Analysis} Layers are matched by relative depth: layer $L$ of Gemma's $26$ corresponds to layer $\operatorname{round}(64L/26)$, giving $12, 25, 37, 49$ for Gemma's $5, 10, 15, 20$ and $0, 2, 5, 7, 10, 12$ for the temperature layers $0$--$5$. Temperature moves use the same $18$ words, three ladders and $49$ moves as Section~\ref{sec:temperature}. Nulls are drawn exactly as for Gemma: $5{,}000$ permutations for temperature and analogies.

\begin{table}[H]
\small
\centering
\begin{tabular}{@{}lccc@{}}
\toprule
 & Gemma & OLMo & Qwen \\
\midrule
Temperature, $^\circ$C & 9/9 & 9/9 & 9/9 \\
Temperature, $^\circ$F & 9/9 & 9/9 & 9/9 \\
Importance is size & 4/4 & 4/4 & 4/4 \\
Argument is war & 4/4 & 4/4 & 4/4 \\
Ideas are food & 4/4 & 2/4 & 3/4 \\
Professional frames$^\ddagger$ & 2 & 2--3 & 2 \\
\bottomrule
\end{tabular}
\caption{Every analysis in all three models (Gemma-2-2B, OLMo-3.1-32B, Qwen3-32B): the number of depth-matched layers at which the move agreement clears its null ($p<0.05$). Temperature is tested at nine layers and everything else at four. $^\ddagger$The number of the $10$ profession pairs that clear, as a range over layers. (Appendix~\ref{app:models})}
\label{tab:models_replication}
\end{table}

\section{Sensitivity analysis}
\label{app:sensitive}

\begin{figure*}[tp]
    \centering
    \includegraphics[width=0.9\linewidth]{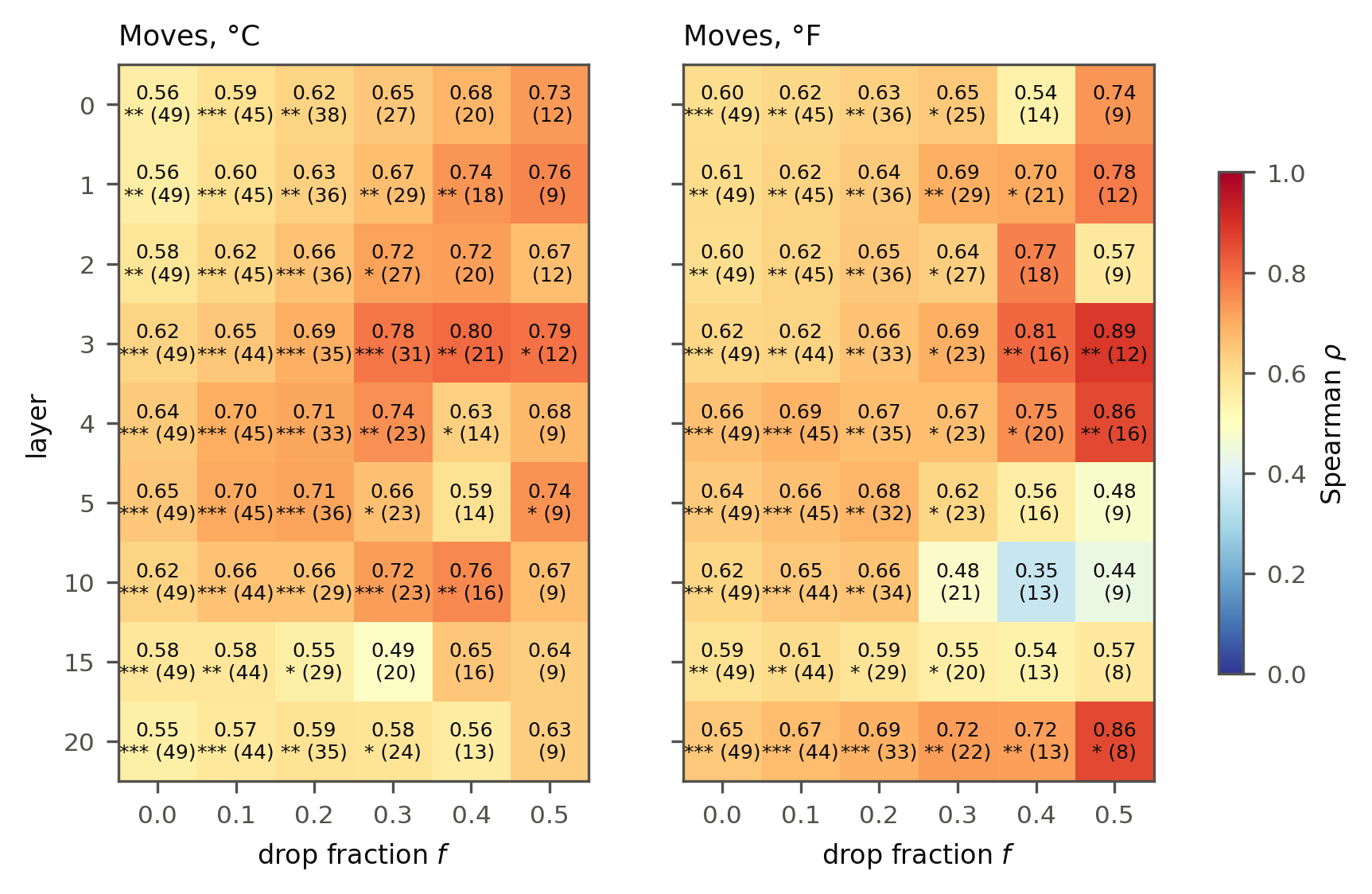}
    \caption{Temperature move agreement under the efficacy sweep, at all nine layers. Each cell is the move RSA after removing the fraction $f$ of water words with the lowest steering score; stars give significance against that cell's own permutation null ($^{***}p<.001$, $^{**}p<.01$, $^{*}p<.05$) and brackets the number of surviving moves. Cells with fewer than $5$ surviving moves are not shown. Removing poorly performing steering vectors does not qualitatively change the findings.}
    \label{fig:temp_sensitivity_sweep}
\end{figure*}
\begin{figure*}[tp]
    \centering
    \includegraphics[width=0.6\linewidth]{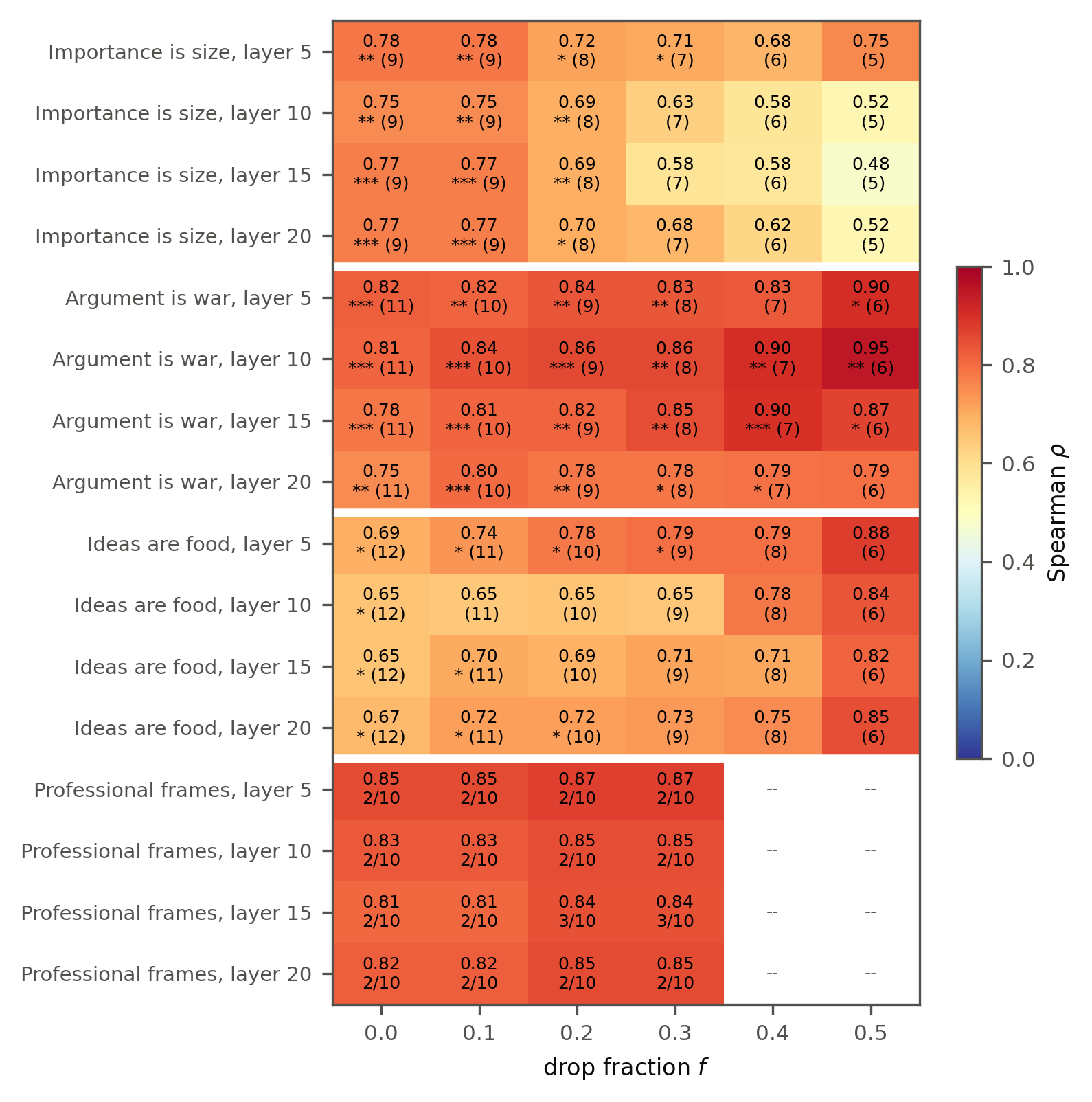}
    \caption{Analogy move agreement under the efficacy sweep. Each cell is the move RSA of Figure~\ref{fig:analogy_rsa} recomputed after removing the fraction $f$ of pairs with the lowest steering score; stars give significance against that cell's own permutation null ($^{***}p<.001$, $^{**}p<.01$, $^{*}p<.05$) and brackets the number of surviving pairs. Frames cells give the mean over the 10 profession pairs and how many of the 10 clear $p<0.05$. \textbf{Argument is war} keeps its agreement at every layer to $f = 0.3$ ($p \le 0.027$) and at three of four layers to $f = 0.5$. \textbf{Importance is size} keeps its agreement to $f = 0.2$. \textbf{Ideas are food} does not survive: at layer $10$ it stops being significant as soon as its one non-steering pair (\emph{swallow}$\to$\emph{accept}, score $0$) is removed, and by $f = 0.4$ it is lost at every layer. \textbf{Professional frames} change little over the range that can be tested. (Appendix~\ref{app:sensitive})}
    \label{fig:analogy_sensitivity_sweep}
\end{figure*}

\textbf{Removing steering vectors that have the lowest LLM judge score does not change the substance of this paper's findings.}

For every vector we use a single number, the LLM-judge Overall score: the harmonic mean with zero gate of the three rubrics (fluency, instruction following, concept relevance), maximised over the 15 steering factors, on a $0$--$2$ scale. It is computed per layer, so a vector may be effective at one depth and not another. Higher is better; 0 means the judge scored the generation 0 at every factor.

We use the judge score rather than the judge-free parse check because it is defined for every concept domain, including the water words, which name no number and so admit no parse-based criterion.

Let the item set of an analysis be $x_1 \dots x_n$ with efficacy scores $s_1 \dots s_n$ at the layer in question. For a drop fraction $f \in \{0, 0.1, 0.2, 0.3, 0.4, 0.5\}$ we remove the $\lfloor fn \rfloor$ items with the smallest $s$, recompute the analysis on the survivors, and recompute its permutation null on the surviving item set. Ties are broken by a fixed item order (nominal temperature, then concept id, for temperature; position in the mapping for analogies), so the sequence is deterministic and nested: the set dropped at $f$ is a subset of the set dropped at any larger $f$. We report each analysis's headline statistic as a function of $f$.

\paragraph{Temperature} An item depends on (at least) two vectors and is only as trustworthy as the weaker of the two, so we score it by the minimum over the pair,
\begin{equation*}
\begin{split}
  s_i = \min\bigl(&\text{Overall}(\text{word}_i),\\
                  &\text{Overall}(\text{numeric level at } t_i)\bigr),
\end{split}
\end{equation*}and drop on that. Removing an item removes the water word and with it every move having that
word as an endpoint. For Fahrenheit the numeric half of the pair is the level at $\tfrac{9}{5}t_i + 32$, so the two units
have different pair scores and different drop orders.

Because the pooled move test permutes words, its resolution falls as words are removed: with $n_w$ surviving words the permutation group has $n_w!$ elements, and the three ladders ($6$, $6$ and $7$ words) begin to lose members. At $f = 0.5$ only 9 of 18 words remain. We report the move analysis over the whole sweep but do not interpret it past $f \approx 0.3$.

\paragraph{Analogies} Items are the pairs of each mapping ($9$, $11$, $12$ and $6$ for size, war, food and profession), not single vectors. As with temperature, a pair enters the analysis through two vectors, so we score it by the weaker of the two,
$$ s_i \;=\; \min\bigl(\text{Overall}(\text{source}_i),\; \text{Overall}(\text{target}_i)\bigr), $$
with ties broken by position in the mapping. Removing a pair removes both vectors and every move ending at either. The professional frames are swept within each of their 10 profession pairs, a role being scored by the minimum over its two professions; with only six roles, $f = 0.4$ and $0.5$ leave four and three, too few to test, and are not reported. At the smallest item sets the null runs out of relabellings: with $n$ surviving items there are only $n!$, so the attainable $p$ is about $1/720 \approx 0.0014$ at $n = 6$ and $1/120 \approx 0.008$ at $n = 5$.

\paragraph{Nulls} Every permutation null is recomputed on the surviving item set, with the same draw counts as the main text. The corresponding $p$-floors are unchanged by the sweep. Where a statistic already sits on its floor at $f=0$, the sweep cannot show it improving.

\subsection{Other models}
\label{app:sensitive-models}

We repeat the sweep in OLMo and Qwen. The sweep runs at every depth where the model was steered. As for Gemma, $f = 0$ reproduces each model's results in Table~\ref{tab:models_replication}. For a compact comparison we summarise each sweep by the largest drop fraction up to which the result remains significant at every step (Table~\ref{tab:models_sweep}).

\begin{table*}[tp]
\small
\centering
\setlength{\tabcolsep}{4pt}
\begin{tabular}{@{}lcccccccc@{}}
\toprule
 & \multicolumn{3}{c}{Depth $\approx 0.19$} & \multicolumn{3}{c}{Depth $\approx 0.38$} & \multicolumn{2}{c}{Depth $\approx 0.58$} \\
\cmidrule(lr){2-4}\cmidrule(lr){5-7}\cmidrule(l){8-9}
 & G & O & Q & G & O & Q & G & O \\
\midrule
Temperature, $^\circ$C & 0.3 & 0.4 & 0.5 & 0.4 & 0.4 & 0.5 & 0.2 & 0.5 \\
Temperature, $^\circ$F & 0.3 & 0.3 & 0.5 & 0.2 & 0.5 & 0.4 & 0.3 & 0.4 \\
Importance is size & 0.3 & 0.2 & 0.3 & 0.2 & 0.3 & 0.2 & 0.2 & 0.4 \\
Argument is war & 0.3 & 0.4 & 0.5 & 0.5 & 0.4 & 0.4 & 0.5 & 0.5 \\
Ideas are food & 0.3 & -- & 0.0 & 0.0 & 0.2 & 0.0 & 0.1 & 0.2 \\
\bottomrule
\end{tabular}
\caption{The largest drop fraction $f$ up to which each move result stays significant ($p<0.05$) at every step of the sweep, at matched relative depth (G = Gemma-2-2B, O = OLMo-3.1-32B, Q = Qwen3-32B). $0.5$ means the result survives the whole sweep; $0.0$ that it is lost at the first drop; -- that it is not significant even at $f=0$. As in Figure~\ref{fig:temp_sensitivity_sweep}, a temperature cell needs at least $5$ surviving moves. (Appendix~\ref{app:sensitive-models})}
\label{tab:models_sweep}
\end{table*}

The substantive results are consistent across models. \textsc{ideas are food} is marginal in all three models. In OLMo it is not significant at the shallowest depth even before any are dropped, and elsewhere they survive at most to $f = 0.2$. In Qwen they are lost at the first drop. At matched item counts ($9$--$10$ pairs), its $z$ is $-0.3$ to $2.7$ across models and depths, against $2.5$ to $4.2$ for \textsc{argument is war}.

\section{Sparse-autoencoder features}
\label{app:sae}

\begin{figure*}[tp]
  \centering
  \includegraphics[width=0.85\linewidth]{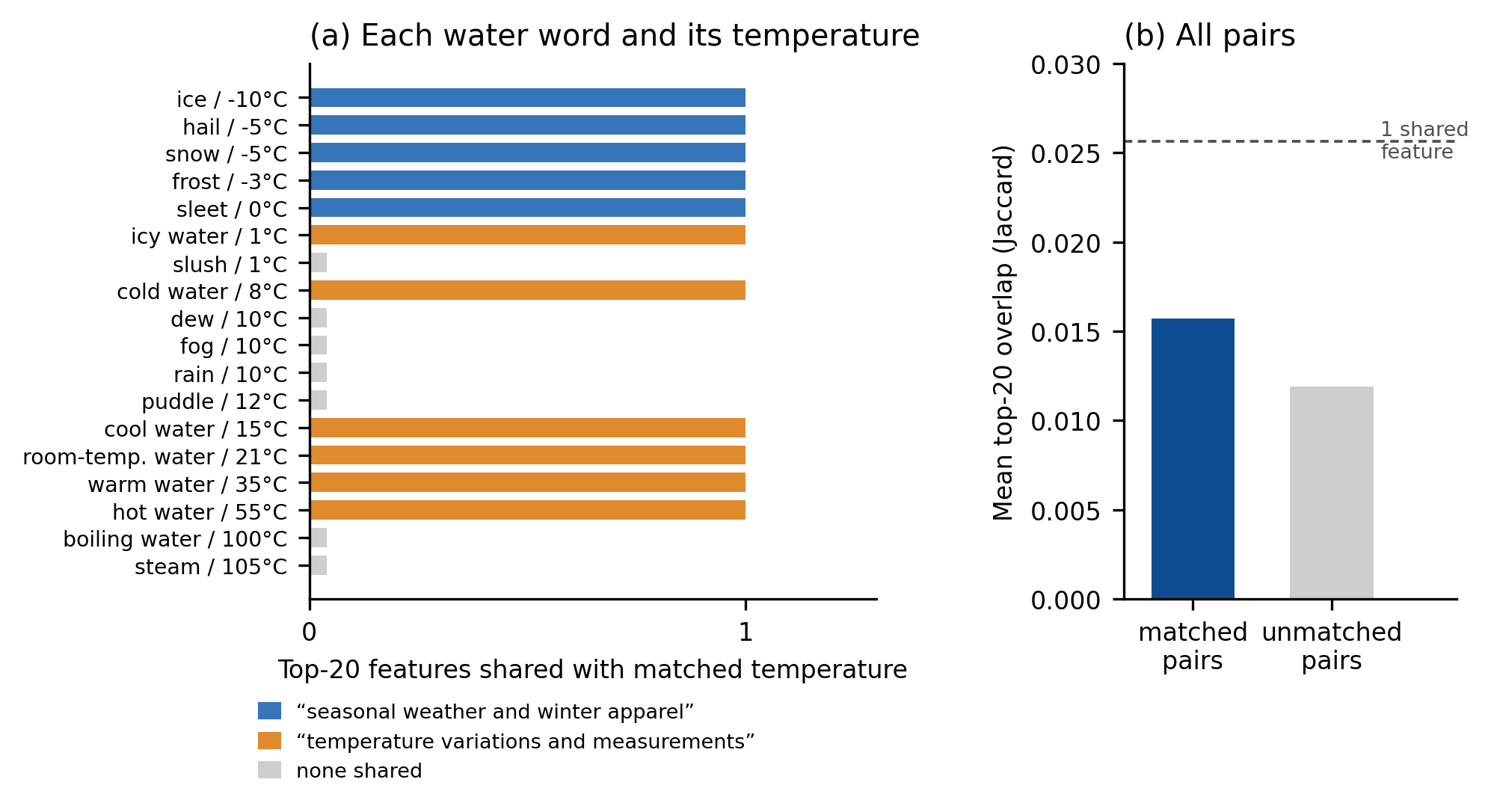}
  \caption{Shared SAE features between water words and numeric temperatures (GemmaScope $16$k, layer $10$). \textbf{(a)}~For each water word, the number of its $20$ strongest features that also appear among the $20$ strongest features of its matched Celsius temperature, coloured by the shared feature. \textbf{(b)}~Mean overlap (Jaccard) for matched pairs and for unmatched pairs. The dashed line marks the overlap of a pair that shares a single feature.}
  \label{fig:sae}
\end{figure*}

\paragraph{Method} We ran \texttt{gemma-2-2b-it} over the same training text used to build every steering vector and encoded the residual stream at layer $10$ with the GemmaScope $16$k-feature SAE. For each concept we then computed the difference of means in \emph{feature space}: the mean SAE activation on its positive examples minus the mean on its negative examples, pooled over tokens exactly as for the steering vectors.

\paragraph{Corresponding concepts do not share features} For each concept (encoded as the difference of SAE features) we took the $20$ most active features, and compared these sets across the two domains (Figure~\ref{fig:sae}b). These are the features that are strongly active on positive text and not active on negative text. \emph{Boiling water} and $100\,^{\circ}$C share none of them, and neither do \emph{steam} and $105\,^{\circ}$C. Where a water word does share a feature with its temperature, it is the same generic feature for a whole group: \emph{ice}, \emph{hail}, \emph{snow}, \emph{frost} and \emph{sleet} each share one ``seasonal weather and winter apparel'' feature with their temperatures, and six of the seven adjectival words share one ``temperature variations and measurements'' feature. Across all pairs, the overlap between corresponding concepts (mean Jaccard similarity of $0.016$ for water and Celsius) is barely above that between non-corresponding concepts ($0.012$). SAEs would not have detected that the model considers these concept clouds related. We think this is because the model encodes the relationship in second-order correspondence rather than a shared direction.

\section{Positions}
\label{app:positions}

The main text compares moves. The same comparison can be made on the steering vectors themselves, which we call \emph{positions}. Within each domain we compute the cosine distance between every pair of concepts,
\[
  P_A[i,j] \;=\; 1 - \cos\bigl(v^{A}_{i},\, v^{A}_{j}\bigr),
\]
and likewise $P_B$, and measure agreement as Spearman's $\rho$ between their upper-triangle entries. The pairing of the two domains and the permutation null are the same as for moves (Section~\ref{sec:method}).
 
Positions keep all of a concept's semantics, while a move cancels what two concepts share. Comparing the two shows how much of the agreement lies in the change we expect, such as temperature, rather than in other semantics.

\begin{table}[H]
\centering
\setlength{\tabcolsep}{3.5pt}
\small
\begin{tabular}{@{}lrrrcrrr@{}}
\toprule
 & \multicolumn{3}{c}{$^{\circ}$C} & & \multicolumn{3}{c}{$^{\circ}$F} \\
\cmidrule(r){2-4}\cmidrule(l){6-8}
Layer & $\rho$ & $z$ & $p$ & & $\rho$ & $z$ & $p$ \\
\midrule
0 & $+0.33$ & $3.5$ & $0.0004$ &  & $+0.33$ & $3.5$ & $0.0012$ \\
1 & $+0.32$ & $3.5$ & $0.0008$ &  & $+0.32$ & $3.4$ & $0.0018$ \\
2 & $+0.38$ & $4.0$ & $0.0002$ &  & $+0.32$ & $3.2$ & $0.0020$ \\
3 & $+0.40$ & $4.3$ & $0.0002$ &  & $+0.38$ & $3.7$ & $0.0002$ \\
4 & $+0.45$ & $4.7$ & $0.0002$ &  & $+0.49$ & $4.8$ & $0.0002$ \\
5 & $+0.43$ & $4.6$ & $0.0002$ &  & $+0.48$ & $4.8$ & $0.0002$ \\
10 & $+0.33$ & $3.4$ & $0.0026$ &  & $+0.48$ & $4.6$ & $0.0002$ \\
15 & $+0.24$ & $2.7$ & $0.0100$ &  & $+0.34$ & $3.4$ & $0.0008$ \\
20 & $+0.30$ & $3.3$ & $0.0040$ &  & $+0.45$ & $4.7$ & $0.0002$ \\
\bottomrule
\end{tabular}
\caption{Positions (Temperature), all layers. RSA between the two domains' position matrices, over $18$ words and $\binom{18}{2}=153$ pairs. $z$ is the observed $\rho$ in standard deviations of the label-permutation null, which is centred at $0$ (mean $|\rho_{\text{null}}| < 0.003$, s.d.\ $\approx 0.09$ at every layer). Every cell clears its null. The null is not exactly normal, so $z$ is an approximation.}
\label{tab:app_node_temp_rsa_layers}
\end{table}

\begin{table}[H]
\small
\centering
\begin{tabular}{@{}lccc@{}}
\toprule
 & Gemma & OLMo & Qwen \\
\midrule
Temperature, $^\circ$C & 9/9 & 9/9 & 9/9 \\
Temperature, $^\circ$F & 9/9 & 8/9 & 8/9 \\
Importance is size & 4/4 & 4/4 & 4/4 \\
Argument is war & 0/4 & 0/4 & 2/4 \\
Ideas are food & 0/4 & 0/4 & 0/4 \\
Professional frames$^\ddagger$ & 3--4 & 3--5 & 3--5 \\
\bottomrule
\end{tabular}
\caption{Positions in all three models (Gemma-2-2B, OLMo-3.1-32B, Qwen3-32B): the number of depth-matched layers at which position agreement clears its null ($p<0.05$); Table~\ref{tab:models_replication} gives the same count for moves. $^\ddagger$The number of the $10$ profession pairs that clear, as a range over layers.}
\label{tab:app_positions_models}
\end{table}

\paragraph{Temperature} Positions agree above chance at every layer, in both units (Table~\ref{tab:app_node_temp_rsa_layers}). 
 
\paragraph{Analogies} Positions recover \textsc{importance is size} at every layer, but not \textsc{argument is war} or \textsc{ideas are food}. The larger models give the same result, except that \textsc{argument is war} clears its null at two of four layers in Qwen (Table~\ref{tab:app_positions_models}). More profession pairs clear their null with positions ($3$--$5$ of $10$) than with moves ($2$--$3$).

Using moves, which cancel shared structure, helps detect representational alignment. The simplest explanation is that the alignment is concentrated in the semantics we are interested in.

\section{Model size and budget}
We used open-source models that were $\approx2$ billion parameters (\texttt{gemma-2-2b-it}), and $\approx32$ billion parameters (\texttt{OLMo-3.1-32B-Instruct} and \texttt{Qwen3-32B}). We used closed-source models of unknown size (\texttt{gemini-3.5-flash}, \texttt{gemini-3.1-flash-lite}).

We did not record exact GPU usage but estimate $\approx$2 days of A100 usage.

We did not record exact API usage but estimate our Gemini usage costs $\approx \$300$.

\end{document}